\documentclass[10pt,twocolumn,letterpaper]{article}

\usepackage[pagenumbers]{iccv} % To force page numbers, e.g. for an arXiv version

\definecolor{iccvblue}{rgb}{0.21,0.49,0.74}
\usepackage[accsupp]{axessibility}
\usepackage[pagebackref,breaklinks,colorlinks,allcolors=iccvblue]{hyperref}
\usepackage{indentfirst}  
\usepackage{bbm}
\usepackage{adjustbox}
\usepackage{setspace}
\usepackage{multirow} % 
\usepackage{colortbl} % 
\usepackage{bm}

\def\paperID{4255} % *** Enter the Paper ID here
\def\confName{ICCV}
\def\confYear{2025}
\normalfont
\title{A Hidden Stumbling Block in Generalized Category Discovery: Distracted Attention}

\author{
Qiyu Xu$^{1,3}$ \quad Zhanxuan Hu$^{1}$$^{\dagger}$ \quad Yu Duan$^{2}$ \quad Ercheng Pei$^{3}$ \quad Yonghang Tai$^{1}$$^{\dagger}$\\
$^{1}$Yunnan Normal University, 
$^{2}$Xidian University \\
$^{3}$Xi’an University of Posts and Telecommunications \\
\tt\small{ \{graceafleve, zhanxuanhu, duanyuee\}@gmail.com }\\
\tt\small{ercheng.pei@xupt.edu.cn, taiyonghang@126.com} \\ 
}

\begin{document}
\maketitle
\begin{abstract}

Generalized Category Discovery (GCD) aims to classify unlabeled data from both known and unknown categories by leveraging knowledge from labeled known categories. While existing methods have made notable progress, they often overlook a hidden stumbling block in GCD: distracted attention. Specifically, when processing unlabeled data, models tend to focus not only on key objects in the image but also on task-irrelevant background regions, leading to suboptimal feature extraction. To remove this stumbling block, we propose Attention Focusing (AF), an adaptive mechanism designed to sharpen the model's focus by pruning non-informative tokens. AF consists of two simple yet effective components: Token Importance Measurement (TIME) and Token Adaptive Pruning (TAP), working in a cascade. TIME quantifies token importance across multiple scales, while TAP prunes non-informative tokens by utilizing the multi-scale importance scores provided by TIME. AF is a lightweight, plug-and-play module that integrates seamlessly into existing GCD methods with minimal computational overhead. When incorporated into one prominent GCD method, SimGCD, AF achieves up to $15.4\%$ performance improvement over the baseline with minimal computational overhead. The implementation code is provided in:\url{https://github.com/Afleve/AFGCD}.
\end{abstract}   
% \vspace{-0.5cm}
\section{Introduction}
\label{sec:intro}

\renewcommand{\thefootnote}{\fnsymbol{footnote}} 
\footnotetext[2]{Corresponding Authors}

\begin{figure}[t]  
    \centering  
    \includegraphics[width=0.45\textwidth]{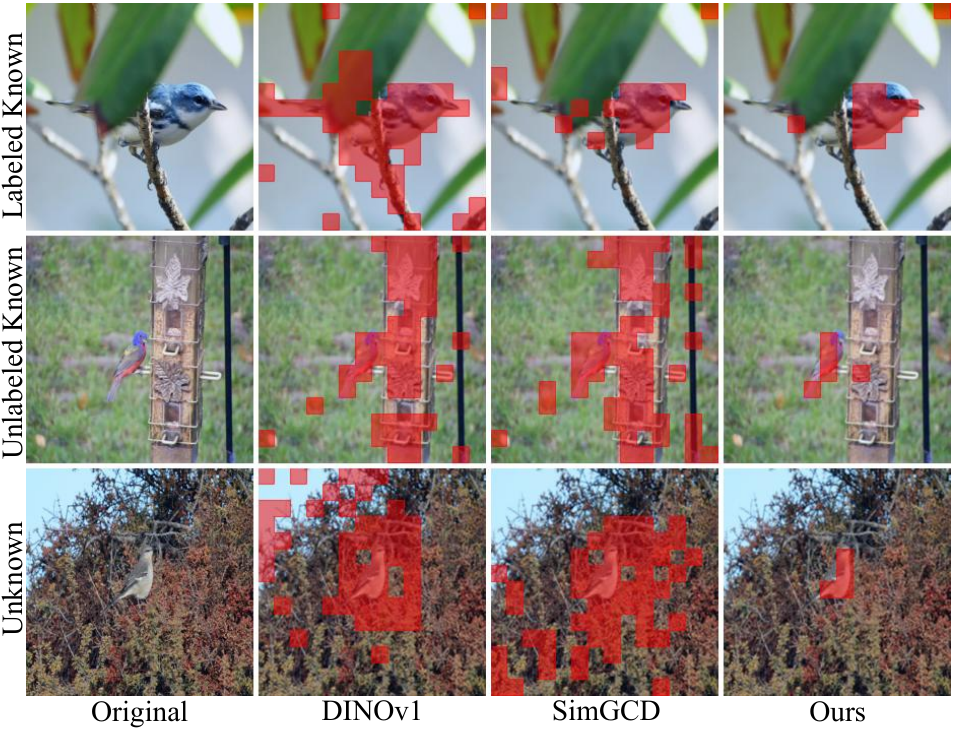}  
    \caption{The masks obtained by thresholding the self-attention maps to retain 70\% of the total mass. DINOv1 and SimGCD demonstrated substantial \emph{distracted attention} on the unlabeled data, meaning it not only focuses on key objects within the image but also on task-irrelevant background regions. In contrast, our method effectively refines the model's focus. More visualization results and analyses can be found in \textbf{Appendix C.1}.}  
    \label{fig:attnMap}
    % \vspace{-0.6cm}
\end{figure}

The rapid advancement of deep learning has led to significant breakthroughs in object recognition, yet many real-world applications demand more than merely classifying data into pre-defined categories. In scenarios such as autonomous driving and medical imaging, models must be capable of discovering and learning from unseen classes. Generalized Category Discovery (GCD) addresses this challenge by leveraging knowledge from a set of labeled known categories to classify unlabeled data that may contain both known and unknown categories. 

Most existing GCD methods follow a standardized learning paradigm: 1) employing a pre-trained Vision Transformer (ViT) as the foundational feature extraction backbone and 2) constructing task-specific GCD heads through the [CLS] token embeddings produced by the backbone. Despite notable progress, they often overlook a hidden stumbling block: \emph{distracted attention}. Specifically, when processing unlabeled data, models tend to distribute their focus not only on key objects but also on irrelevant background regions.  To investigate this, we examine one prominent GCD method, SimGCD~\cite{simGCD}, on a challenging dataset, \texttt{CUB}~\cite{welinder2010caltech}. As illustrated in Figure \ref{fig:attnMap}, visualization of self-attention scores in the final block of ViT shows that while the [CLS] tokens for labeled data consistently concentrate on foreground objects, those for unlabeled data, particularly from unknown categories, exhibit pronounced associations with background regions. This unintended capture of extraneous information degrades the quality of feature representations and, consequently, model performance.

% Our empirical analysis reveals that 

We hypothesize that \emph{distracted attention} arises partly from data augmentation. For labeled data, images within the same class often display varied backgrounds, prompting the model to concentrate on the key objects. In contrast, augmentations applied to unlabeled data typically introduce only minor variations in the background, enabling the model to exploit spurious correlations as shortcuts in unsupervised or self-supervised learning. Based on this assumption, a straightforward solution is to prune task-irrelevant tokens from the input image, ensuring that the model’s decision relies exclusively on tokens pertinent to the key object.

To this end, we propose Attention Focusing (AF), an adaptive mechanism designed to sharpen the model's focus by pruning non-informative tokens. As shown in Figure~\ref{fig:framework}, AF consists of two simple yet effective components: Token Importance Measurement (TIME) and Token Adaptive Pruning (TAP), working in a cascade. In practice, TIME introduces a task-specific query token in each ViT block to quantify token importance across multiple scales. Subsequently, TAP utilizes the multi-scale importance scores generated by TIME to prune non-informative tokens, mitigating the interference from task-irrelevant information. 

Benefiting from its straightforward design, AF is a lightweight, plug-and-play module that integrates seamlessly into existing GCD methods with minimal computational overhead. In this paper, we integrate AF into SimGCD for two primary reasons. First, SimGCD employs an exceptionally simple architecture that effectively combines supervised and self-supervised learning, without introducing overly complex modules. Second, SimGCD has already demonstrated promising results across a wide range of datasets. To evaluate the effectiveness of AF, we extensively test the improved method on seven publicly available GCD datasets. The experimental results reveal that AF significantly boosts the performance of SimGCD, especially on fine-grained datasets with complex background information. Remarkably, these significant performance improvements are achieved with minimal computational overhead. This demonstrates that AF offers a highly efficient enhancement to the existing GCD framework.
The main contributions of this work are summarized as follows:
\begin{enumerate}
    \item \emph{ Novel perspective.} To the best of our knowledge, we are the first to investigate and quantify the harmful effects of \emph{distracted attention} in GCD. This incredible finding provides a new direction toward improving this field.
    \item \emph{ Novel method.} We propose AF, a simple yet effective module that provides the first generic solution for attention correction in GCD through token adaptive pruning.
    \item  \emph{Promising results.} We evaluate the effectiveness and efficiency of AF across different settings. Experimental results demonstrate that AF can significantly improve performance with minimal computational overhead.
\end{enumerate}
\section{Related Work}
\subsection{Generalized Category Discovery}
GCD extends the paradigms of Semi-Supervised Learning (SSL)~\cite{li2021comatch, fini2023semi} and Novel Category Discovery (NCD)~\cite{UNO}, which leverages knowledge of known categories within open-world settings to simultaneously identify both known and unknown classes from unannotated data. Most existing GCD methods can be broadly categorized into: 1) non-parametric methods; and 2) parametric methods.

Non-parametric methods~\cite{2022CVPRgcd, pu2023dynamic, 2023promptcal, rastegar2024learn, CMS2024, GPC} typically involve training a feature extractor followed by the application of clustering techniques, such as semi-supervised K-means++~\cite{2022CVPRgcd}, to obtain the final classification results. For example, GCD~\cite{2022CVPRgcd} introduces a fundamental framework that utilizes traditional supervised and unsupervised contrastive learning to achieve effective representation learning. Similarly, DCCL~\cite{pu2023dynamic} optimizes instance-level and concept-level contrastive objectives through dynamic conception generation and dual-level contrastive learning, exploiting latent relationships among unlabeled samples. Furthermore, GPC~\cite{2023gpc} integrates a Gaussian Mixture Model within an Expectation-Maximization framework to alternate between representation learning and category estimation, SelEx~\cite{SelEx} introduces ‘self-expertise’ to enhance the model’s ability to recognize subtle differences. In addition, PromptCAL~\cite{2023promptcal} utilizes visual prompt tuning to facilitate contrastive affinity learning within a two-stage framework, while CMS~\cite{CMS2024} incorporates Mean Shift clustering into the contrastive learning process to encourage tighter grouping of similar samples.

Parametric methods~\cite{simGCD, wang2024sptnet, get} integrate the optimization of a parametric classifier to directly yield prediction outcomes. For instance, SimGCD~\cite{simGCD} jointly trains a prototype classifier alongside representation learning, establishing a robust baseline for this category of methods. SPTNet~\cite{wang2024sptnet} employs a two-stage framework that alternates between model refinement and prompt learning. Moreover, GET~\cite{get} leverages CLIP to generate semantic prompts for novel classes via text generation, thereby unlocking the potential of multimodal models for addressing the GCD task.

Indeed, most existing GCD methods primarily focus on how to leverage unsupervised or self-supervised learning techniques to enhance model performance on unlabeled data. Despite notable progress, they often overlook a hidden stumbling block: \emph{distracted attention}. Addressing this challenge is the core of this paper. It is worth noting that during the review process, we identified two representative works that also aim to mitigate background interference~\cite{aptgcd,mos}. Nevertheless, our method differs fundamentally in both its underlying motivation and methodological design.
\subsection{High-Resolution Image Recognition}
High-resolution recognition refers to the capability of computer vision systems to accurately identify and analyze objects in images characterized by a high pixel count and intricate details. Managing \emph{distracted attention} is a critical challenge in this context, as the extensive spatial information often leads to inefficient feature extraction and model focus drift.  A widely adopted strategy to address this issue is to partition high-resolution images into smaller patches, thereby increasing the relative proportion of key targets within each patch. For instance, IPS~\cite{bergner2022iterative} iteratively processes individual patches and selectively retains those most relevant to the specific task. SPHINX~\cite{zhang2023llamaadapter} segments a high-resolution image into a set of low-resolution images and concatenates these with a downsampled version of the original image as the visual input. Monkey~\cite{li2023monkey} employs a sliding window approach combined with a visual resampling mechanism to enhance image resolution, thereby improving content comprehension while reducing computational overhead. Furthermore, LLaVA-UHD~\cite{guo2024llava-uhd} ensures both efficiency and fidelity in image processing by optimizing slice computation and scoring functions, effectively minimizing resolution variations. 
On one hand, these methods are specifically designed for supervised learning scenarios and cannot be directly applied to GCD tasks without significant modifications. On the other hand, we process the original images directly, achieving greater efficiency while preserving accuracy.

\subsection{Token Pruning}
Another issue closely related to this work is token pruning, which aims to enhance computational efficiency and reduce redundancy by selectively removing task-irrelevant patches while preserving most of the original image information. EVit~\cite{liang2022evit} leverages the attention values between the [CLS] token and patch tokens in ViT to select the most informative patches. SPVit~\cite{kong2022spvit} and SVit~\cite{SVit} propose retaining pruned tokens from upper layers for subsequent use, rather than discarding them entirely. PS-ViT (T2T)~\cite{tang2022patch} adopts a reverse approach by selecting tokens for pruning based on the final output features. ToMe~\cite{bolya2022tome} reduces the computational workload by merging tokens with high key similarity. While these methods have achieved notable advancements in improving inference efficiency, they often result in varying degrees of performance degradation. In the context of the GCD task, however, model accuracy is of paramount importance. Additionally, many methods rely on the [CLS] token for pruning, but in the GCD task, the [CLS] token for unlabeled data tends to be of lower quality, making it susceptible to introducing misleading information~(see \textbf{Appendix C.3}).
The method most relevant to ours is Cropr~\cite{bergner2024token}, which prunes a fixed number of tokens at each ViT block. However, we adopted multi-scale adaptive pruning to address the diversity of image backgrounds, achieving better results~(see Section~\ref{sec:fixed_num}). 

\section{Method}
\subsection{Problem Formulation}
Generalized Category Discovery (GCD) addresses the problem of automatically clustering unlabeled data \(\mathcal{D}^{u} = \{(x_{i}, y_{i}^{u}) \in \mathcal{X} \times \mathcal{Y}_{u}\}\) in a partially labeled dataset \(\mathcal{D}^{l} = \{(x_{i}, y_{i}^{l}) \in \mathcal{X} \times \mathcal{Y}_{l}\}\). Here, \(\mathcal{Y}_{l}\) represents the set of known classes, and \(\mathcal{Y}_{u}\) represents the set of all classes, with \(\mathcal{Y}_{l} \subset \mathcal{Y}_{u}\). In different GCD approaches, the number of unknown classes \( |\mathcal{Y}_{u}|\) can be utilized as prior knowledge or estimated through established methods.
\subsection{Overview}
\begin{figure}[t]  
    \centering  
    \includegraphics[width=0.45\textwidth]{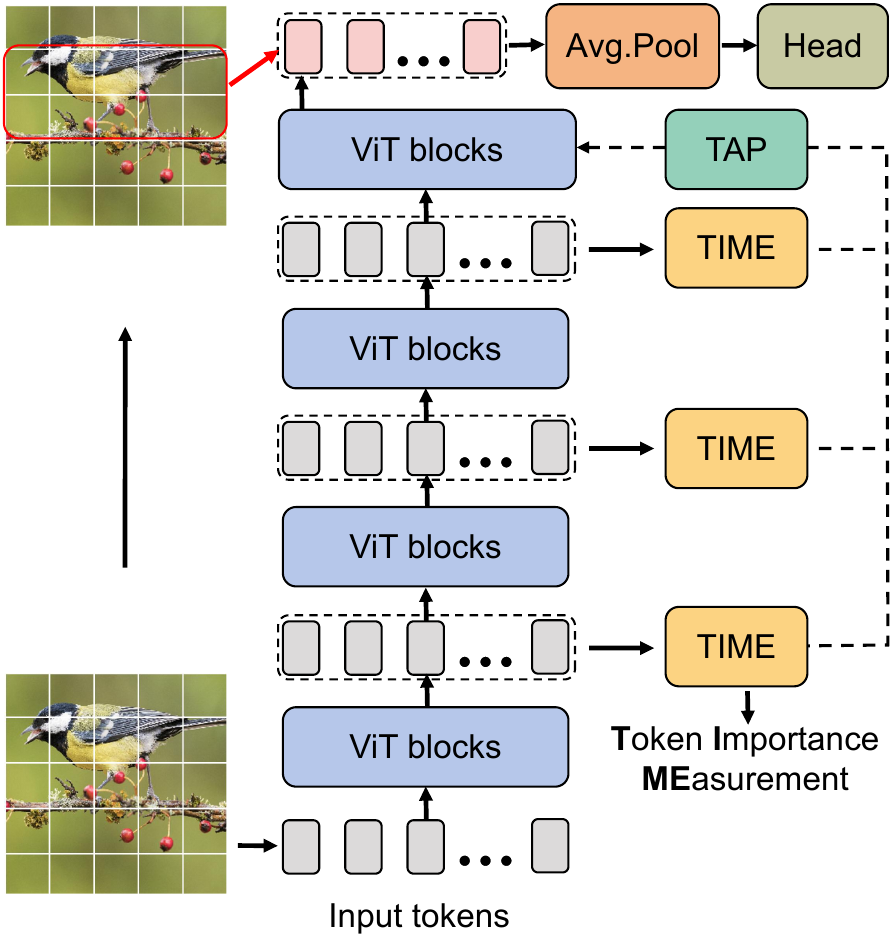}  
    \caption{The pipeline of GCD with our proposed \emph{Attention Focusing(AF)} mechanism. AF consists of two components: Token Importance Measurement (TIME) and Token Adaptive Pruning (TAP), working in a cascade. Here, the '\emph{Head}' can be inherited from any existing GCD model. }  
    \label{fig:framework}  
\end{figure}  
The currently popular GCD methods are primarily based on pre-trained ViT models. Specifically, given an image \(I \in \mathbb{R}^{h \times w \times 3}\), ViT divides it into a sequence of non-overlapping patches, each of size \(P \times P\). This sequence of patches is then flattened and mapped into token embeddings \(\{ \textbf{x}_n \in \mathbb{R}^{1 \times D}, n = 1, 2, 3, ..., N \}\) through a linear projection head, where $N=H \times W, H = h/P , W = w/P$, and \( D \) represents the dimensionality of the embedding space. 
After appending an additional [CLS] token to the patch tokens, the resulting token sequence \( \mathbf{X} \in \mathbb{R}^{(N+1) \times D} \) is passed sequentially through all transformer blocks. For simplicity, the batch size \(B\) and block number $l$ are omitted from the description. Ultimately, the [CLS] token produced by the backbone network is passed into the task-specific GCD head. As illustrated in Figure~\ref{fig:attnMap}, while the [CLS] tokens for labeled data consistently focus on foreground objects, those for unlabeled data, especially from unknown categories, show strong associations with background regions. This unintended capture of extraneous information degrades the quality of feature representations and, consequently, the performance of the GCD model.

To this end, we propose integrating a novel AF mechanism into the existing GCD model. As illustrated in Figure~\ref{fig:framework}, the AF mechanism consists of two simple yet effective components: Token Importance Measurement (TIME) (Section~\ref{sec:time}) and Token Adaptive Pruning (TAP) (Section~\ref{sec:tap}), which operate in a cascade. In practice, the TIME module is inserted into every block of the ViT, except for the last one. Each TIME module outputs a score vector that reflects the importance of each patch token. The TAP module then aggregates these multi-scale scores to prune the non-informative tokens. Finally, the remaining tokens are processed with average pooling and then used as input to the \textit{Head}. It is important to note that the \textit{Head} can be inherited from any existing GCD method. In this work, our primary experiment is based on SimGCD~\cite{simGCD}, a representative GCD method. Additionally, we integrate the AF mechanism into three representative methods, CMS~\cite{CMS2024}, GET~\cite{get}, and SelEx~\cite{SelEx}, to demonstrate its generalizability (see Section~\ref{sec:CMS}). Next, we will provide a detailed description of TIME and TAP, while further details on SimGCD can be found in the \textbf{Appendix A}.

\subsection{Token Importance Measurement}
\label{sec:time}

\begin{figure}[t]
    \centering 
    \includegraphics[width=0.32\textwidth]{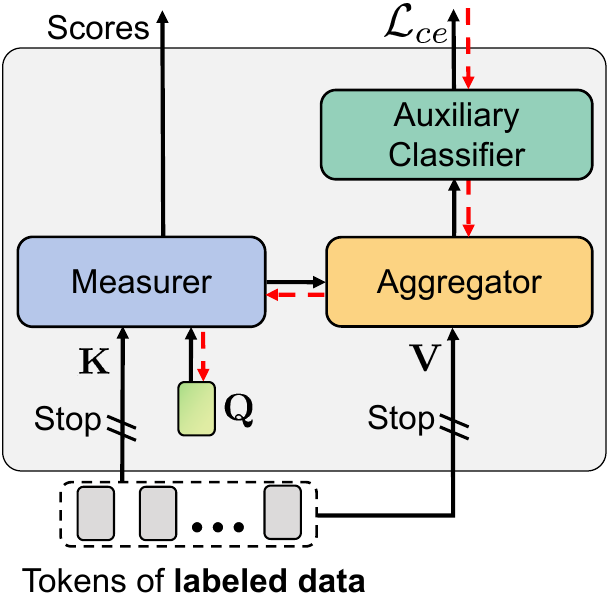}  
    \caption{The internal pipeline of TIME. The red dashed lines represent the gradient propagation paths from the auxiliary classifier to the optimization of \( \mathbf{Q} \). Besides, TIME is trained using only labeled data, but it works on both labeled and unlabeled data. }
    \label{fig:time}  
\end{figure}  

As shown in Figure~\ref{fig:time}, TIME is trained exclusively on labeled data but is capable of generalizing to the entire training set. Specifically, given an image, TIME takes its tokens as input and produces a score vector \( \mathbf{s} \in \mathbb{R}^{1 \times (N+1)} \), revealing the informativeness of the input tokens. Specifically, each TIME module consists of three key components: a \emph{Measurer}, an \emph{Aggregator}, and an \emph{Auxiliary classifier}.

The \textit{Measurer} assigns the score vector \( \boldsymbol{s} \in \mathbb{R}^{1 \times (N+1)} \) to each token by performing cross-attention between the tokens and a learnable query vector \( \mathbf{Q} \).
Specifically, the input tokens \( \mathbf{X} \) are treated as the key matrix \( \mathbf{K} \) and value matrix \( \mathbf{V} \). The query vector \( \mathbf{Q} \) is then used to query \( \mathbf{K} \), yielding attention results for each token. The scores between the query vector and the key matrix are computed as follows:
\begin{equation}
\mathbf{s}(\mathbf{Q}, \mathbf{K}) = \frac{\mathbf{Q} \mathbf{K}^T}{\sqrt{D}},
\end{equation}
where \( \sqrt{D} \) is a scaling factor to stabilize the attention values.
To ensure the informativeness scores \( \mathbf{s} \) are properly utilized, the \emph{Aggregator} leverages these scores to obtain an initial image representation. Specifically, the aggregated representation \( \mathbf{r} \) is computed as:
\begin{equation}
\mathbf{r} = \text{Softmax}(\mathbf{s}) \mathbf{V}.
\end{equation}
Furthermore, to increase the capacity of the \emph{Aggregator}, we follow~\cite{bergner2024token} and incorporate a transformer block's Feed-Forward Network (FFN), which includes LayerNorm (LN) and an MLP with a residual connection. Mathematically, 
\begin{equation}
\mathbf{r}' = \text{MLP}(\text{LayerNorm}(\mathbf{r})) + \mathbf{r}.
\end{equation}
Next, the resulting representation \( \mathbf{r}' \) is passed through the \emph{Auxiliary classifier}, producing a probability output \( \mathbf{p} \in \mathbb{R}^{1 \times |\mathcal{Y}_l|} \), where \( |\mathcal{Y}_l| \) is the number of possible classes for labeled data. TIME is trained using a cross-entropy loss:
\begin{equation}
\mathcal{L}_{ce} = -\sum_{k=1}^{|\mathcal{Y}_l|} y^k \log p^k,
\end{equation}
where \( y^k \) represents the ground truth label and \( p^k \) is the predicted probability.

In practice, the \emph{Auxiliary classifier} aids in classifying labeled data, guiding the \emph{Aggregator} to focus on the most informative features of the image that are crucial for classification. As training progresses, the query vector \( \mathbf{Q} \) dynamically adjusts the score vector \( \mathbf{s} \), assigning progressively higher importance to tokens with greater informativeness. This adaptive mechanism enables the model to prioritize the most relevant tokens for the task, improving its ability to capture critical information for accurate classification. Generally, unlabeled data and labeled data often share similar stylistic characteristics. Therefore, we hypothesize that the query vector \( \mathbf{Q} \), learned from labeled data, generalizes well and can effectively assess the importance of patch tokens even in the case of unlabeled data.

Additionally, we apply a stop-gradient to isolate the \emph{Auxiliary classifier} from the backbone, ensuring that conflicting gradients do not affect the encoder. During testing, the \emph{Auxiliary classifier} is discarded, and only the query vector \( \mathbf{Q} \) is retained to process the test samples. This reduces computational overhead while maintaining the model's capacity to evaluate token importance effectively.

%Besides, we apply a stop-gradient to isolate the auxiliary classifier from the backbone. This approach prevents the encoder from being influenced by conflicting gradients from the auxiliary classifier. Moreover, the auxiliary classifier is discarded during testing, and only the query vector  is retained to process the test samples. This results in minimal computational overhead while preserving the model's ability to assess token importance effectively.

\subsection{Token Adaptive Pruning}
\label{sec:tap}

The score vectors obtained from different TIME blocks represent the importance of patch tokens across different scales. TAP leverages these multi-scale importance scores to prune the input patch tokens. Specifically, given a set of score vectors \( \{ \mathbf{s}_l \in \mathbb{R}^{1 \times (N+1)} \}_{l=1}^{L-1} \), where \( L \) denotes the number of ViT blocks, the multi-scale importance of patch tokens is computed as follows:
\begin{equation}
\mathbf{s}^{m} = \frac{1}{L-1} \sum_{l=1}^{L-1} \text{Softmax}(\hat{\mathbf{s}}_l),
\label{eq:s}
\end{equation}
where \( \hat{\mathbf{s}}_l \in \mathbb{R}^{1 \times N} \) represents a score vector that excludes the value associated with the [CLS] token. This exclusion is crucial because the [CLS] token aggregates high-level semantic information about the image, making it a meaningful token in itself. Next, for the patch tokens \( \mathbf{X} = (\mathbf{x}_1, \mathbf{x}_2, \ldots, \mathbf{x}_N) \), we prune the less informative tokens by applying an adaptive threshold \( \tau \). Formally, we define the pruned patch tokens \( \mathbf{X}_p \) as:
\begin{equation}
\mathbf{X}_p = \{ \mathbf{x}_i \mid i = 1, 2, \ldots, t, \sum_{i=1}^{N} s_i^m \leq \tau \},
\label{eq:filter}
\end{equation}
where \( s_i^m \) is the \( i \)-th element of the multi-scale importance score vector \( \mathbf{s}^{m} \), and the indices \( i = 1, 2, \ldots, N \) are sorted in increasing order of \( s_i^m \). The pruned patch tokens \( \mathbf{X}_p \) represent redundant information associated with task-irrelevant regions in the image. The remaining token sequence, \( \mathbf{X}_r \), consisting of the residual patch tokens and the [CLS] token, is then passed through the final ViT block. Finally, the output token representations are processed using average pooling to form the final image representation, which is subsequently input into the GCD \textit{Head}. The overall loss function of our improved method is: 
\begin{equation}
    \mathcal{L}=\mathcal{L}_{gcd}+\lambda\sum_{l=1}^{L-1}\mathcal{L}_{ce}^l\,,
\end{equation}
where $\mathcal{L}_{gcd}$ denotes the loss function of the selected GCD baseline model, $\lambda$ is a balancing parameter.
\subsection{Discussion}
During the training process of GCD, each instance is typically augmented with two distinct views, raising an important question: \emph{Should we adopt single-view TAP or multi-view TAP?} The former applies TAP to only one of these views, while the latter applies TAP to both augmented views simultaneously. In this work, we opt for single-view TAP for two main reasons. First, TAP can be seen as a form of non-regular image cropping augmentation, where single-view TAP is particularly effective in helping the model focus on key objects of interest. By pruning unnecessary tokens in a single view, the model can retain critical information, improving its ability to extract meaningful features from the complex image. Second, multi-view TAP effectively forces the model to train without the interference of background information across both views. Although this may appear beneficial in theory by reducing noise, it can inadvertently hinder the model's ability to generalize~(as shown in  \textbf{Appendix C.2}).

%And, TAP is applied to only one of these views. To further assess the potential benefits of a more comprehensive approach, we experimented with multi-view TAP, where TAP is applied to both augmented views simultaneously. As shown in Table~\ref{res:crop}, while multi-view TAP does offer some performance improvements, it also leads to a noticeable degradation in comparison to single-view TAP. We believe this can be attributed to two primary factors.  

\section{Experiments}
\label{sec:experiments}
 
\subsection{Experimental Setup}
\noindent\textbf{Dataset.} In this study, we primarily incorporate AF into SimGCD~\cite{simGCD} and evaluate the effectiveness using three challenging fine-grained datasets from the Semantic Shift Benchmark~\cite{SSB2021}: \texttt{CUB}~\cite{welinder2010caltech}, \texttt{Stanford Cars}~\cite{krause20133d}, and \texttt{FGVC-Aircraft}~\cite{maji2013fine}. Additionally, we apply our method to three more generic classification datasets, namely \texttt{CIFAR10/100}~\cite{krizhevsky2009learning} and \texttt{ImageNet-100}~\cite{deng2009imagenet}, as well as the large-scale fine-grained dataset \texttt{Herbarium-19}~\cite{tan2019herbarium}. As discussed in the \textbf{Appendix B.1}, the former often includes complex background information, while the latter exhibits relatively minimal background interference. To ensure the fairness of the experiments, all other settings are kept consistent with SimGCD. More details can be found in the \textbf{Appendix A}. %All experiments were conducted on an NVIDIA GeForce RTX 4090 GPU.

\vspace{4pt}
\noindent\textbf{Evaluation.}
Following established practice~\cite{simGCD}, we utilize clustering accuracy (ACC) to evaluate the model performance. Prior to comparing the ground truth with the predicted labels, we employ the Hungarian algorithm~\cite{hungarian} to align the labels of the \emph{Unknown} category, followed by calculating the accuracy (ACC) using \(
\frac{1}{M} \sum_{i = 1}^{M} \mathbbm{1}(y_{i}^* = p(\hat{y}_{i}))
\) where \(M\) = \(|D_U|\), and \(p\) denotes the optimal permutation.

For clarity and convenience, the accuracy metrics are reported for {\textquotesingle \emph{All}\textquotesingle\ unlabeled data, along with the subsets corresponding to known and unknown classes, labeled as \textquotesingle \emph{Old}\textquotesingle\ and \textquotesingle \emph{New}\textquotesingle\ in the tables, respectively.

\subsection{Main Results}
%We conduct experiments to compare with the state-of-the-art methods (RankStates~\cite{han2021autonovel}, UNO+~\cite{UNO}, ORCA~\cite{2019iccvncd}, GCD~\cite{2022CVPRgcd}, DCCL~\cite{2023CVPRDCC}, GPC~\cite{GPC}, SimGCD~\cite{simGCD}, PIM~\cite{PIM2023}, InfoSieve~\cite{rastegar2024learn}, CMS~\cite{CMS2024}, SPTNet~\cite{wang2024sptnet}) in GCD. The detailed results are as follows.

\vspace{4pt}
\noindent\textbf{Evaluation on challenging fine-grained datasets.}
Table~\ref{res:fine_grained} presents a comparison between SimGCD and several state-of-the-art methods on three challenging fine-grained datasets, where \textquotesingle $\triangle$\textquotesingle\ denotes the performance improvements over the baseline model, SimGCD. Clearly, SimGCD serves as a robust baseline model, achieving competitive results in the vast majority of settings, despite its simple network architecture. Comparing with SimGCD+AF, we observe that the AF module significantly enhances the model’s performance, underscoring its effectiveness in addressing the \emph{distracted attention} issue in SimGCD. Compared to other state-of-the-art methods, SimGCD+AF consistently achieves the best or near-best performance across various datasets. On the \texttt{CUB} dataset, the performance of InfoSieve and CMS is comparable to that of SimGCD+AF. However, SimGCD+AF demonstrates a clear advantage on the other two datasets, particularly on \texttt{Stanford Cars}, where the performance improvement on \textquotesingle \emph{All}\textquotesingle\ reaches up to 10.1\%. While SPTNet and SimGCD+AF perform similarly on \texttt{FGVC-Aircraft}, SPTNet's performance on \texttt{Stanford Cars} is notably weaker than that of SimGCD+AF. Additionally, SPTNet employs an alternating training strategy, resulting in a higher computational cost compared to SimGCD+AF. Both MOS and AptGCD also focus on mitigating the interference of background information and achieve results comparable to SimGCD+AF. However, AF is relatively simpler in module design and does not rely on any external models.
%Although  demonstrate comparable performance to our approach, they are specifically tailored for fine-grained datasets, which may limit their generalizability across broader domains.  AptGCD also aims to enhance generalization through attention refinement, its design introduces substantially higher model complexity due to the integration of multiple visual prompt modules and auxiliary objectives, thereby limiting its scalability and practicality in real-world scenarios. The performance of AF is fundamentally bounded by the baseline capabilities of SimGCD. In extended experiments, we demonstrate that integrating the AF module with more advanced GCD methods yields substantially superior performance.
% \vspace{-0.2cm}
\begin{table}[ht]
    % \hspace{-0cm}
    \footnotesize
    \centering
\begin{tabular}{l p{0.3cm} p{0.3cm} p{0.3cm} p{0.3cm} p{0.3cm} p{0.3cm} p{0.3cm} p{0.3cm} p{0.3cm}}
\toprule
\multicolumn{1}{l}{\multirow{2}{*}{Datasets}} & \multicolumn{3}{c}{CUB} & \multicolumn{3}{c}{Stanford Cars} & \multicolumn{3}{c}{FGVC-Aircraft}  \\
       & All & {Old} & {New} & All & {Old} & {New} & All & {Old} & {New}   \\ 
       \midrule
         RankStats~\cite{han2021autonovel}  & 33.3 & 51.6 & 24.2 & 28.3 & 61.8 & 12.1 & 26.9 & 36.4 & 22.2  \\ 
         UNO+~\cite{UNO}  &35.1 & 49.0 & 28.1 & 35.5 & 70.5 & 18.6 & 40.3 & 56.4 & 32.2  \\ 
         ORCA~\cite{2019iccvncd}  &35.3 & 45.6 & 30.2 & 23.5 & 50.1 & 10.7 & 22.0 & 31.8 & 17.1  \\ 
         GCD~\cite{2022CVPRgcd}  &51.3 & 56.6 & 48.7 & 39.0 & 57.6 & 29.9 & 45.0 & 41.1 & 46.9  \\ 
         DCCL~\cite{2023CVPRDCC}  &63.5 & 60.8 & 64.9 & 43.1 & 55.7 & 36.2 & - & - & -   \\ 
         GPC~\cite{GPC} & 55.4 & 58.2 & 53.1 & 42.8 & 59.2 & 32.8 & 46.3 & 42.5 & 47.9  \\
         PIM~\cite{PIM2023}  &62.7 & 75.7 & 56.2 & 43.1 & 66.9 & 31.6 & - & - & -  \\
         InfoSieve~\cite{rastegar2024learn} & {69.4} & \textbf{77.9} & {65.2} & 55.7 & 74.8 & 46.4 & 56.3 & {63.7} & 52.5 \\ 
         CMS~\cite{CMS2024}  &68.2  & \underline{76.5} & 64.0 & 56.9 & 76.1 & 47.6 & 56.0 & 63.4 & 52.3  \\
         % GCA~\cite{GCA} & 68.8 & 73.4 & \textbf{66.6} & 52.0 & 57.1 & \underline{49.5} & 54.4 & \textbf{72.1} & 45.8  \\
         SPTNet~\cite{wang2024sptnet} & 65.8 & 68.8 & 65.1 & {59.0} & {79.2} & {49.3} & {59.3} & 61.8 & {58.1} \\
         %SelEx~\cite{SelEx} & \underline{73.6} & 75.3 & \underline{72.8} & 58.5 & 75.6 & 50.3 & 57.1 & 64.7 & 53.3\\
         AptGCD~\cite{aptgcd} & \textbf{70.3} & 74.3 & \textbf{69.2} & 62.1 & 79.7 & 53.6 & \textbf{61.1} & 65.2 & \textbf{59.0} \\
         MOS~\cite{mos} & \underline{69.6} & 72.3 & \underline{68.2} & \underline{64.6} & \textbf{80.9} & \underline{56.7} & \textbf{61.1} & \underline{66.9} & \underline{58.2} \\
         %GET~\cite{get} & \textbf{77.0} & \textbf{78.1} & \textbf{76.4} & \textbf{78.5} & \textbf{86.8} & \textbf{74.5} & 58.9 & 59.6 & \underline{58.5} \\
         \midrule
         SimGCD~\cite{simGCD}  &60.3 & 65.6 & 57.7 & 53.8 & 71.9 & 45.0 & 54.2 & 59.1 & 51.8 \\
         SimGCD+AF & {69.0} & 74.3 & {66.3} & \textbf{67.0} & \underline{80.7} & \textbf{60.4} & \underline{59.4} & \textbf{68.1} & {55.0} \\ 
            $\triangle$ & \textcolor{green}{+8.7} & \textcolor{green}{+8.7} & \textcolor{green}{+8.6} & \textcolor{green}{+13.2} & \textcolor{green}{+8.8} & \textcolor{green}{+15.4} & \textcolor{green}{+5.2} & \textcolor{green}{+9.0} & \textcolor{green}{+3.2} \\
         \bottomrule 
         
    \end{tabular}
    \caption{Comparison with several state-of-the-art methods on fine-grained datasets. The best results are highlighted in \textbf{bold}, and the second-best results are highlighted in \underline{underline}. \textquotesingle $\triangle$\textquotesingle\ refers to the performance improvement compared to SimGCD~\cite{simGCD}.}
    \label{res:fine_grained}
\end{table}

\vspace{4pt}
\noindent\textbf{Evaluation on generic datasets.}
Table~\ref{res:generic} presents the results on generic datasets. We observed that the improvement brought by AF on these datasets is less pronounced than on the fine-grained datasets. We attribute this to two main factors. First, the SimGCD model has already achieved excellent performance on these datasets, such as nearly 100\% accuracy on \texttt{CIFAR-10}. Second, the backgrounds of these datasets are relatively simple, leading to minimal interference. For example, on \texttt{CIFAR-100}, due to the lack of complex backgrounds, AF even resulted in a performance decrease for the new classes. In contrast, for \texttt{ImageNet100}, a dataset with more complex backgrounds, AF provided a more noticeable performance improvement. Compared to other methods, SimGCD+AF also achieves competitive results, but it typically involves lower computational cost.

\begin{table}[t]
    \hspace{-0.6cm}
    \footnotesize
    \centering
\begin{tabular}{l p{0.3cm} p{0.3cm} p{0.3cm} p{0.3cm} p{0.3cm} p{0.45cm} p{0.3cm} p{0.3cm} p{0.3cm}}
\toprule
\multicolumn{1}{l}{\multirow{2}{*}{Datasets}} & \multicolumn{3}{c}{CIFAR10} & \multicolumn{3}{c}{CIFAR100} & \multicolumn{3}{c}{ImageNet-100}  \\
       & All & {Old} & {New} & All & {Old} & {New} & All & {Old} & {New}   \\ 
       \midrule
         RankStats~\cite{han2021autonovel}  &46.8 & 19.2 & 60.5  & 58.2 & 77.6 & 19.3 & 37.1 & 61.6 & 24.8  \\ 
         UNO+~\cite{UNO}   &68.6 & \textbf{98.3} & 53.8 & 69.5 & 80.6 & 47.2 & 70.3 & 95.0 & 57.9  \\ 
         ORCA~\cite{2019iccvncd}  &  96.9 & 95.1 & 97.8 & 69.0 & 77.4 & 52.0 & 73.5 & 92.6 & 63.9   \\ 
         GCD~\cite{2022CVPRgcd}  & 91.5 & 97.9 & 88.2 & 73.0 & 76.2 & 66.5 & 74.1 & 89.8 & 66.3   \\ 
         DCCL~\cite{2023CVPRDCC}  &  96.3 & 96.5 & 96.9 & 75.3 & 76.8 & 70.2 & 80.5 & 90.5 & 76.2    \\ 
         GPC~\cite{GPC}   & 92.2 & \underline{98.2} & 89.1 & 77.9 & \underline{85.0} & 63.0 & 76.9 & 94.3 & 71.0  \\
         PIM~\cite{PIM2023}  & 94.7 & 97.4 & 93.3 & 78.3 & 84.2 & 66.5 & 83.1 & {95.3} & 77.0  \\
         InfoSieve~\cite{rastegar2024learn} & 94.8 & {97.7} & 93.4 & 78.3 & 82.2 & 70.5 & 80.5 & 93.8 & 73.8 \\ 
         CMS~\cite{CMS2024}  & - & - & - &  \underline{82.3} & \textbf{85.7} & 75.5 & {84.7} & \textbf{95.6} & 79.2  \\
         % GCA~\cite{GCA} & 95.5 & 95.9 & 95.2 & \textbf{82.4} & \underline{85.6} & 75.9 & 82.8 & 94.1 & 77.1  \\
         SPTNet~\cite{wang2024sptnet} & \underline{97.3} & 95.0 & {98.6} & 81.3 & 84.3 & 75.6 & \underline{85.4} & 93.2 & \underline{81.4} \\
         %SelEx~\cite{SelEx} & 95.9 & {98.1} & 94.8 & \underline{82.3} & 85.3 & 76.3 & 83.1 & 93.6 & 77.8 \\
         AptGCD~\cite{aptgcd} & \underline{97.3} & 95.8 & \underline{98.7} & \textbf{82.8} & 81.8 & \textbf{85.5} & \textbf{87.8} & \underline{95.4} & \textbf{84.3} \\
         %GET~\cite{get} & 97.2 & 94.6 & 98.5 & 82.1 & \underline{85.5} & 75.5 & \textbf{91.7} & \textbf{95.7} & \textbf{89.7} \\
         \midrule
         SimGCD~\cite{simGCD}  & 97.1 & 95.1 & 98.1 & 80.1 & 81.2 & \underline{77.8} & 83.0 & 93.1 & 77.9 \\
         SimGCD+AF & \textbf{97.8} & 95.9 & \textbf{98.8} & {82.2} & {85.0} & {76.5} & \underline{85.4} & 94.6 & {80.8} \\ 
            $\triangle$ & \textcolor{green}{+0.7} & \textcolor{green}{+0.8} & \textcolor{green}{+0.7} & \textcolor{green}{+2.1} & \textcolor{green}{+3.8} & \textcolor{red}{-1.3} & \textcolor{green}{+2.4} & \textcolor{green}{+1.5} & \textcolor{green}{+2.9}   \\
         \bottomrule 
         
    \end{tabular}
    \caption{Comparison with several state-of-the-art methods on three generic datasets.}
    \label{res:generic}
\end{table}
\vspace{4pt}

\noindent\textbf{Evaluation on more challenging datasets.}
Compared to the above three fine-grained datasets, \texttt{Herbarium-19} has a simpler background, and as a result, the performance gain brought by AF is also relatively limited. This highlights a limitation of our method AF: while it effectively suppresses interference from background information, it does not significantly improve the model's ability to extract information from the key objects themselves.

\begin{table}[t]
    \footnotesize
    \centering
\begin{tabular}{l p{0.8cm} p{0.8cm} p{0.8cm} p{0.8cm} p{0.8cm} p{0.8cm}}
\toprule
\multicolumn{1}{l}{\multirow{2}{*}{Datasets}} & \multicolumn{3}{c}{Herbarium-19} \\
       & All & {Old} & {New}  \\ 
       \midrule
         GCD~\cite{2022CVPRgcd}  &   35.4 & 51.0 & 27.0 \\ 
         PIM~\cite{PIM2023}  &  42.3 & 56.1 & 34.8 \\
         InfoSieve~\cite{rastegar2024learn} & 41.0 & 55.4 & 33.2 \\ 
         CMS~\cite{CMS2024}  & 36.4 & 54.9 & 26.4 \\
         SPTNet~\cite{wang2024sptnet} & 43.4 & \underline{58.7} & 35.2\\
         \midrule
         SimGCD~\cite{simGCD}  & \underline{44.0} & 58.0 & \underline{36.4}  \\
         SimGCD+AF & \textbf{45.5} & \textbf{59.0} & \textbf{38.3} \\
         $\triangle$ & \textcolor{green}{+1.5} & \textcolor{green}{+1.0} & \textcolor{green}{+1.9}\\
         \bottomrule 
         
    \end{tabular}
    \caption{Comparison with several state-of-the-art methods on Herbarium-19.}
    \label{res:challenge}
    \vspace{-0.5cm}
\end{table}

\begin{figure*}[htbp]
    \centering
        \includegraphics[height=2.8cm]{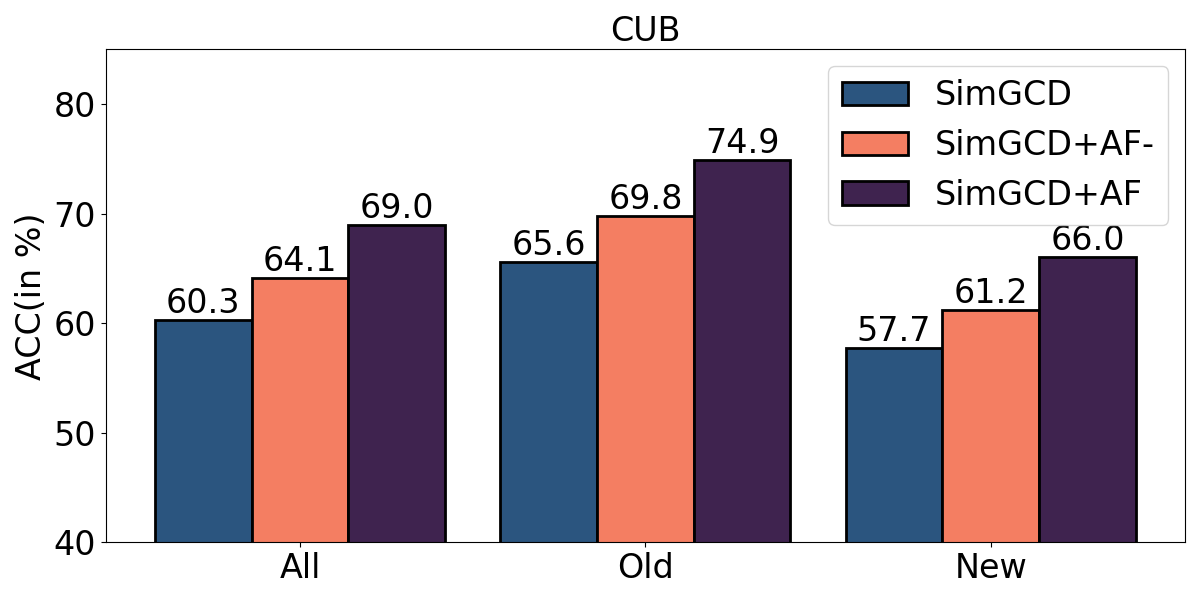}
        \includegraphics[height=2.8cm]{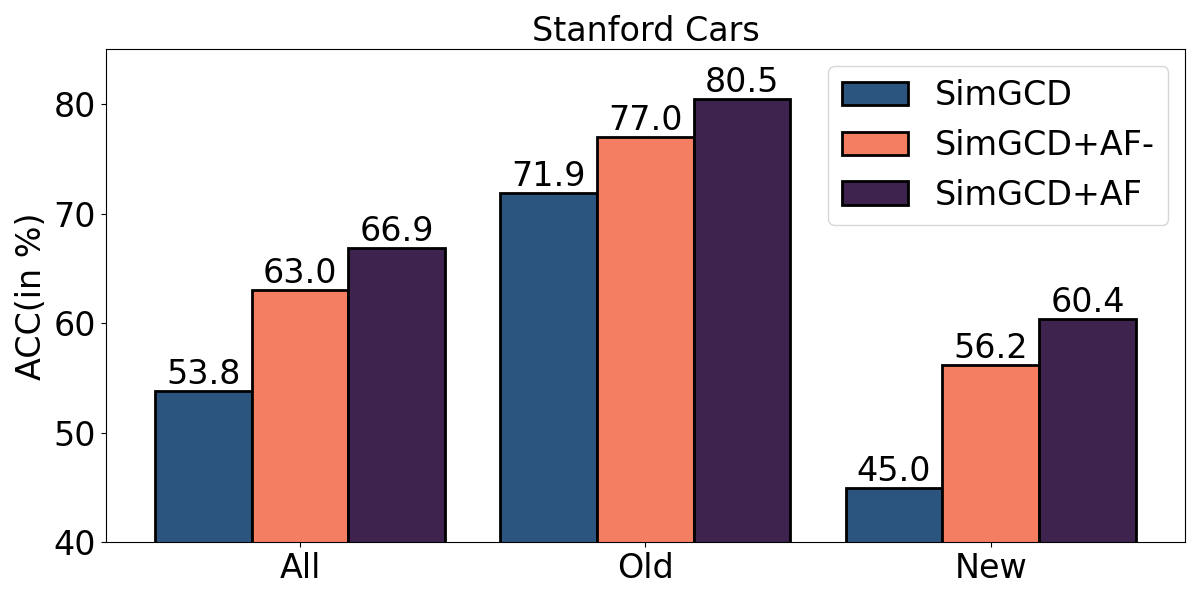}
        \includegraphics[height=2.8cm]{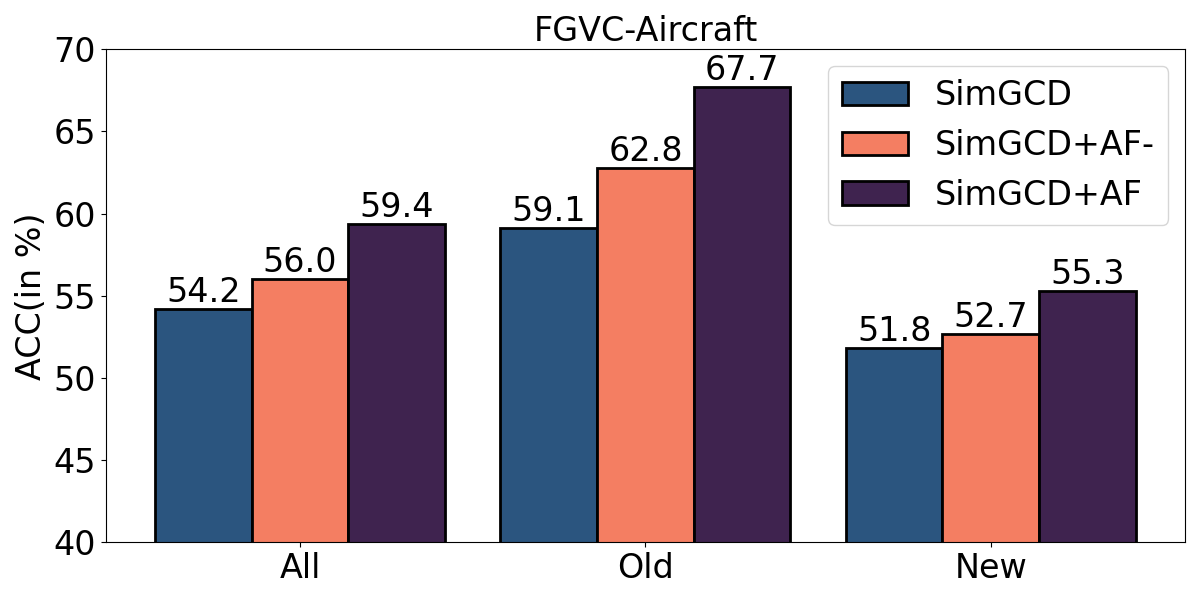}
    \caption{Investigation of Multi-scale token importance measurement. "SimGCD+AF-" refers to a setting where only the query from the penultimate block is used as the basis for token pruning within TAP.}
    \label{fig:multi}
\end{figure*}

\begin{figure*}[h]  
    \centering  
    \includegraphics[width=0.95\textwidth]{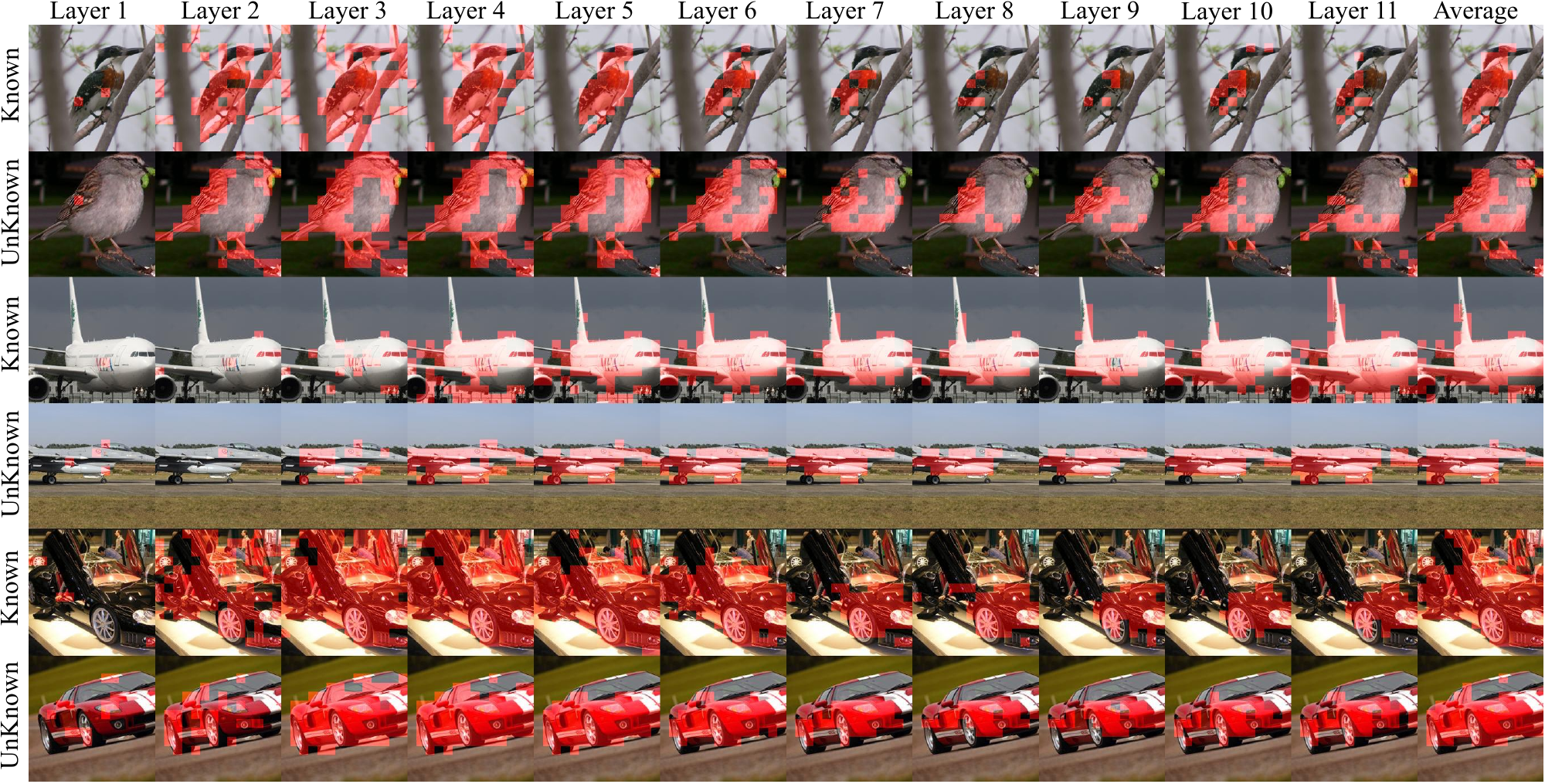}  
    \caption{The results of token pruning using query vectors from each layer. Specifically, the last column illustrates the multi-scale token importance measurement used in AF.}  
    \label{fig:diff_layer}  
\end{figure*} 
\subsection{Discussion on the design of AF}

\vspace{4pt}
\noindent\textbf{Is AF effective for other GCD models?} As mentioned above, AF is a plug-and-play module that can be seamlessly integrated into existing GCD methods without requiring extensive modifications. To further assess the generalizability and effectiveness of AF, we incorporated it into three additional GCD methods, CMS~\cite{CMS2024}, SelEx~\cite{SelEx}, and GET~\cite{get}. The results, as displayed in the Table~\ref{res:cms}, reveal a substantial improvement in performance across various datasets, with particularly notable enhancements observed in the \texttt{Stanford Cars} and \texttt{FGVC-Aircraft} datasets. These findings provide strong evidence of AF's ability to significantly boost the performance of baseline models, highlighting its broad applicability and compatibility with different GCD approaches.
\label{sec:CMS}
\begin{table}[h]
    \scriptsize 
    \centering
    % \hspace{-0.9cm}
\begin{tabular}{l p{0.3cm} p{0.3cm} p{0.3cm} p{0.3cm} p{0.3cm} p{0.3cm} p{0.3cm} p{0.3cm} p{0.3cm}}
\toprule
\multicolumn{1}{l}{\multirow{2}{*}{Datasets}} & \multicolumn{3}{c}{CUB} & \multicolumn{3}{c}{Stanford Cars} & \multicolumn{3}{c}{FGVC-Aircraft}  \\
       & All & Old & {New} & All & {Old} & {New} & All & {Old} & {New}   \\ 
       \midrule
       % GCD & 51.3 & 56.6 & 48.7 & 39.0 & 57.6 & 29.9 & 35.4 & 51.0 & 27.0\\
       % \rowcolor{lightgray}
       % \textbf{GCD+AF} & 55.3 & 54.6 & 55.6 & 47.6 & 51.7 & 45.7 & 48.7 & 47.7 & 49.2\\
       % \midrule
       % $\text{SimGCD}_{clip}$ & 71.3 & 75.7 & 69.1 & 71.2 & 87.0 & 63.6 & 55.7 & 61.0 & 53.0  \\
       % \textbf{$\text{SimGCD}_{clip}$+AF} & 78.0 & 81.1 & 76.4 & 78.1 & 86.0 & 74.3 & 57.2 & 66.0 & 52.7 \\
       % \midrule
         CMS & 67.3 & 75.6 & 63.1 & 53.1 & 73.0 & 43.5 & 54.2 & 63.2 & 49.8  \\
         \rowcolor{lightgray}
         \textbf{CMS+AF} & 68.2 & 75.9 & 64.3 & 61.8 & 76.3 & 54.8 & 57.5 & 62.7 & 54.9 \\ 
       \midrule
       SelEx & 73.4 & 73.9 & 73.2 & 58.9 & 78.6 & 49.4 & 57.2 & 66.3 & 52.6 \\
       % \rowcolor{lightgray}
       \textbf{SelEx+AF} & 79.2 & 76.3 & 80.6 & 61.2 & 80.1 & 52.0 & 62.8 & 66.5 & 60.9 \\
       \midrule
         GET &  75.2 & 77.9 & 73.9 & 78.3 & 86.0 & 74.6 & 57.4 & 59.6 & 54.7 \\
        % \rowcolor{lightgray}
        \textbf{GET+AF} & 77.3 & 77.1 & 77.4 & 81.5 & 90.6 & 77.1 & 59.5 & 67.0 & 55.8 \\ 
        
         \bottomrule 
    \end{tabular}
    \caption{Results of incorporating AF into three additional methods: CMS~\cite{CMS2024}, SelEx~\cite{SelEx} GET~\cite{get}. Notably, CMS did not perform mean shift clustering during testing.} 
    \label{res:cms}
\end{table}
%\vspace{4pt}

\noindent\textbf{Is multi-scale token importance measurement necessary?}
In this work, TAP prunes less informative tokens by aggregating importance scores across multiple scales. Figure~\ref{fig:diff_layer} illustrates the selected patches at different ViT blocks. As shown, the patches selected by the model vary significantly across different layers, primarily due to the differences in the feature scales at each layer. This variability underscores the need for a multi-scale approach, as it enables the model to capture a broader range of key object information, leading to a more robust and comprehensive understanding of the image. Besides, we explored using only the query from the penultimate block as the basis for token pruning in TAP. While this approach still results in some performance improvements for the baseline model SimGCD, as depicted in the Figure~\ref{fig:multi}, the model's performance degrades substantially when compared to SimGCD+AF. This result highlights the necessity of integrating multi-scale token importance measurement.

\vspace{4pt}
\noindent\textbf{Learn queries from only labeled data or all training data?} To empower the \emph{queries} with the capability of selectively attending to informative image tokens, the learnable \emph{queries} in AF are exclusively trained on labeled data. This design choice is motivated by two critical considerations. First, in the absence of supervisory signals, the model struggles to accurately identify and focus on the true key objects within unlabeled images, as the background clutter and irrelevant regions may dominate the feature representation. Second, and more importantly, the self-distillation loss, which is commonly employed in unlabeled data, can inadvertently introduce noise and bias into the learning process of \emph{queries}, thereby deteriorating their ability to distinguish between informative and non-informative patches. This phenomenon is empirically validated in Table~\ref{res:num_classes}, where we observe that training the \emph{queries} on the entire dataset (including both labeled and unlabeled samples) results in a substantial performance drop across all benchmarks. This degradation underscores the importance of leveraging clean, supervised signals for learning robust and discriminative \emph{queries} that can effectively guide the model's attention towards task-relevant tokens. 

\begin{table}[h]
    \footnotesize
    \centering
\begin{tabular}{l p{0.3cm} p{0.3cm} p{0.3cm} p{0.3cm} p{0.3cm} p{0.3cm} p{0.3cm} p{0.3cm} p{0.3cm}}
\toprule
\multicolumn{1}{l}{\multirow{2}{*}{Datasets}} & \multicolumn{3}{c}{CUB} & \multicolumn{3}{c}{Stanford Cars} & \multicolumn{3}{c}{FGVC-Aircraft}  \\
       & All & Old & {New} & All & {Old} & {New} & All & {Old} & {New}   \\ 
       \midrule
       SimGCD & 60.1 & 69.7 & 55.4 & 55.7 & 73.3 & 47.1 & 53.7 & 64.8 & 48.2 \\
         +AF(all)  & 67.4 & 73.9 & 64.1 & 63.0 & 81.5 & 54.1 & 54.6 & 60.5 & 51.6 \\
         \rowcolor{lightgray}
         \textbf{AF} & {69.0} & 74.3 & {66.3} & {67.0} & {80.7} & {60.4} & {59.4} & {68.1} & {55.0} \\ 
         \bottomrule 
    \end{tabular}
    \caption{Investigation of \emph{Query learning}. 'AF(all)' refers to a setting where \emph{Query learning} is based on the entire training dataset.}
    \label{res:num_classes}
\end{table}

\begin{figure}[h]  
    \centering  
    \includegraphics[width=0.3\textwidth]{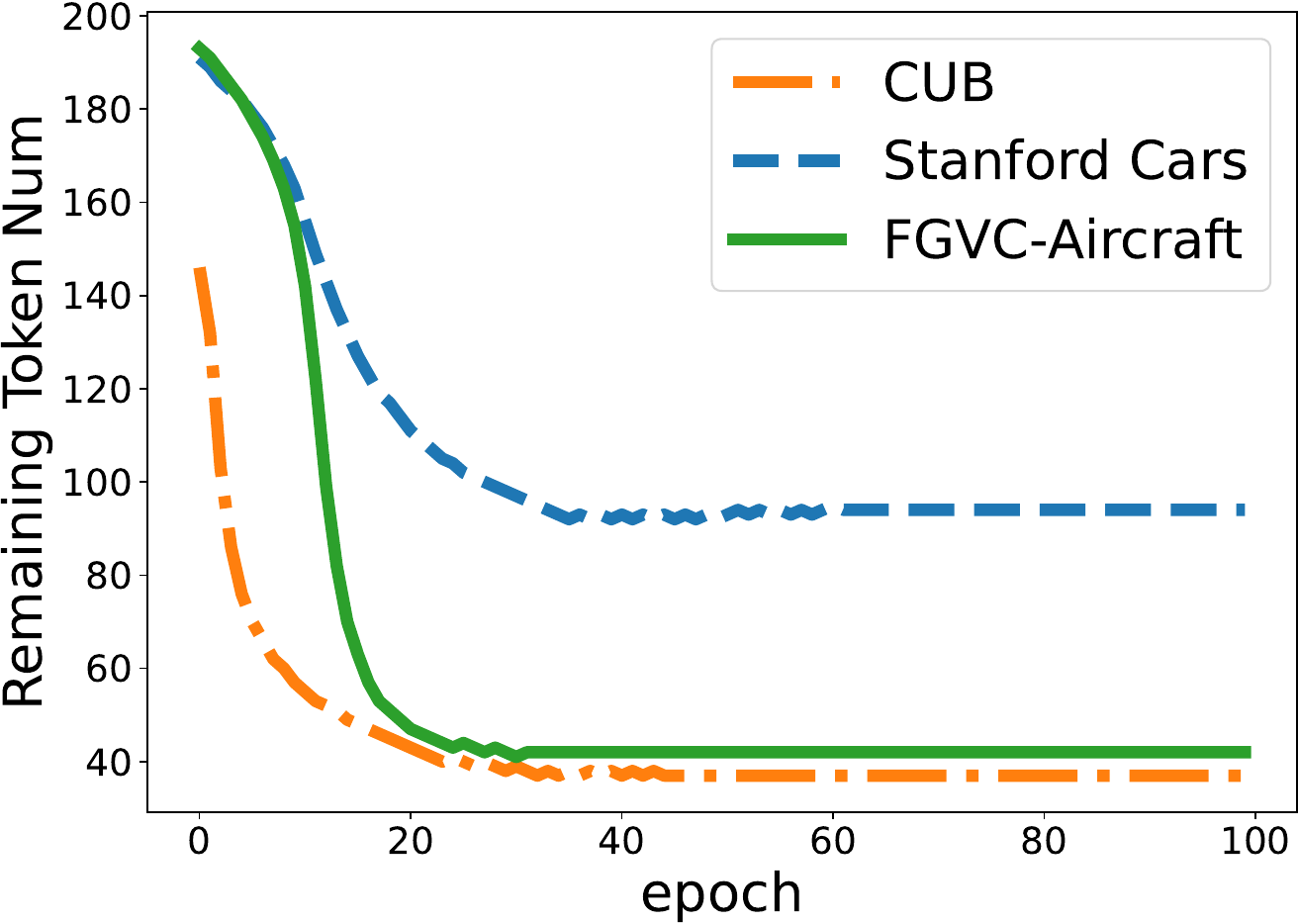}  
    \caption{The dynamic change of the number of retaining patches during the training process.}  
    \label{fig:curve}  
    \vspace{-0.3cm}
\end{figure} 
\noindent\textbf{How important is token adaptive pruning?} Considering the inherent variability in background information across different images, we adopt a token-adaptive pruning strategy in TAP instead of employing a fixed pruning approach. To demonstrate the superiority of TAP, we conduct a comparative experiment using fixed pruning, where a predetermined number of \( k \) patches are uniformly removed from training images. As illustrated in Table~\ref{tab:fixed_num}, while the model's performance exhibits some improvement as the number of removed patches increases within a limited range, it consistently falls short of the performance achieved by TAP. Notably, when \( K = 128 \), the model's performance on the \texttt{Stanford Cars} degrades compared to \( K = 64 \), likely due to the excessive removal of informative patches, which undermines the model's ability to capture essential features. This observation is further corroborated by Figure~\ref{fig:curve}, which reveals that TAP retains a higher proportion of patches on the \texttt{Stanford Cars} dataset compared to \texttt{CUB} and \texttt{FGVC-Aircraft}. These findings underscore the importance of a dynamic, image-specific pruning strategy, as implemented in TAP, to effectively balance the removal of non-informative background patches while preserving critical visual information.
\label{sec:fixed_num}
\begin{table}[h]
    \hspace{-0.6cm}
    \footnotesize
    \centering
\begin{tabular}{l p{0.3cm} p{0.3cm} p{0.3cm} p{0.3cm} p{0.3cm} p{0.3cm} p{0.3cm} p{0.3cm} p{0.3cm}}
\toprule
\multicolumn{1}{l}{\multirow{2}{*}{Datasets}} & \multicolumn{3}{c}{CUB} & \multicolumn{3}{c}{Stanford Cars} & \multicolumn{3}{c}{FGVC-Aircraft}  \\
       & All & Old & {New} & All & {Old} & {New} & All & {Old} & {New}   \\ 
       \midrule
         SimGCD  & 60.1 & 69.7 & 55.4 & 55.7 & 73.3 & 47.1 & 53.7 & 64.8 & 48.2  \\
         $k=16$ & 65.1 & 74.1 & 60.5 & 60.4 & 75.2 & 53.3 &  54.1 & 64.7 & 48.8 \\
         $k=64$ & 67.1 & 72.3 & 64.5 & 63.5 & 79.8 & 55.6 & 54.3 & 61.3 & 50.7 \\
         $k=128$ & 67.0 & 75.0 & 63.0 & 62.4 & 82.8 & 52.6 & 55.5 & 64.9 & 50.7 \\
         \rowcolor{lightgray}
         \textbf{TAP} & {69.0} & 74.3 & {66.3} & {67.0} & {80.7} & {60.4} & {59.4} & {68.1} & {55.0} \\ 
         \bottomrule 
    \end{tabular}
    \caption{Investigation of \emph{Token Adaptive Pruning}. '$k$' refers to a setting where a predetermined number of \( k \) patches are uniformly removed from training images.} 
    \label{tab:fixed_num}
    \vspace{-0.5cm}
\end{table}

% \vspace{4pt}

% \vspace{4pt}
% \noindent\textbf{The impact of TAP on target representation}

\section{Conclusion}
\label{sec:conclusion}
In this work, we introduced AF, a simple yet powerful mechanism designed to address the issue of distracted attention in GCD. By pruning non-informative tokens, AF refines the model’s focus on the key objects in the image, resulting in enhanced performance across both known and unknown categories. Extensive experiments show that when integrated with existing GCD methods, such as SimGCD, AF leads to substantial performance gains while maintaining minimal computational overhead. However, while AF effectively mitigates background interference, it does not significantly improve the model’s ability to extract more discriminative features from the key objects themselves. This limitation points to an avenue for future research: developing methods that can further enhance the model’s ability to focus on the most relevant features of the key objects.
\section*{Acknowledgments}
This work is supported by the National Natural Science Foundation of China~(No.62201453), the Basic Research Project of Yunnan Province~(No.202501CF070004), and the Xingdian Talent Support Program.
{\small
\bibliographystyle{ieeenat_fullname}
\bibliography{reference}
}

\setlength{\parskip}{0pt}
%% TITLE
\maketitlesupplementary
% Supplementary materials table of contents  
\section*{Contents}  
\addcontentsline{toc}{section}{Table of Contents for Supplementary Materials}  

\begin{itemize}[label={}]
    \item \textbf{A. SimGCD} \hfill \pageref{sec:simgcd}  
        % \begin{itemize}[label={}]
        %     \item \textbf{A.1 SimGCD} \hfill \pageref{sec:simgcd}  
        %     \item \textbf{A.2 Implement Details} \hfill       
        %     \pageref{sec:implement}  
        % \end{itemize}  
    \item \textbf{B. Experimental Setup} \hfill \pageref{sec:setup}
        \begin{itemize}[label={}]
            \item \textbf{B.1 The details of datasets} \hfill \pageref{sec:dataset}  
            \item \textbf{B.2 Implementation details} \hfill       
            \pageref{sec:implement}  
        \end{itemize}  
    
    \item \textbf{C. Extended Discussions} \hfill \pageref{sec:discuss}
        \begin{itemize}[label={}]
            % \item \textbf{B.1 The details of datasets and configures} \hfill \pageref{sec:detail} 
            \item \textbf{C.1 The impact of AF on model attention} \hfill \pageref{sec:App_attenMap} 
            \item \textbf{C.2 Single-view TAP or Multi-view TAP?} \hfill \pageref{sec:single-view} 
            \item \textbf{C.3 [CLS] token attention vs. AF} \hfill \pageref{sec:cls_attention} 
            \item \textbf{C.4 The impact of resolution} \hfill \pageref{sec:resolution} 
            \item \textbf{C.5 Class Token or Aggregation Token?} \hfill \pageref{sec:clstoken} 
            \item \textbf{C.6 Computational efficiency of AF} \hfill \pageref{sec:efficiency} 
            \item \textbf{C.7 Parameter analysis} \hfill \pageref{sec:parameter} 
        \end{itemize}  
\end{itemize}  

\appendix

\section{SimGCD} \label{sec:simgcd}
In this work, our primary experiment is based on SimGCD, a representative parametric GCD method that comprises two key components: (1) representation learning and (2) classifier learning. 

\textbf{1)Representation Learning} employs supervised contrastive learning on labeled samples, and self-supervised contrastive learning on all samples. Specifically, given two augmented views $\bm{x}_i$ and $\bm{x'}_i$ of the same image in a batch $B$. The unsupervised contrastive loss is written as:
\begin{equation}  
\mathcal{L}^{u}_\text{rep} = \frac{1}{|B|} \sum_{i \in B} - \log \frac{\exp\left(\bm{z}_i^{\top}  \bm{z}'_i / \tau_u\right)}{\sum^{i \neq n}_i \exp\left(\bm{z}_i^{\top} \bm{z}'_n / \tau_u\right)},  
\end{equation}
where $\bm{z}=g(f(\bm{x}))$ and is $\ell_2$-normalized, $g$ is a MLP projection head, $f$ is the feature backbone, $\tau_u$ is a temperature value. 

The supervised contrastive loss is employed to enhance feature representation by leveraging labeled data to pull samples from the same class closer in the feature space while pushing apart samples from different classes, formally written as:

\begin{equation}
\mathcal{L}^{s}_{\text{rep}} = \frac{1}{|B^l|} \sum_{i \in B^l} \frac{1}{|\mathcal{N}_i|} \sum_{q \in \mathcal{N}_i} -\log \left( \frac{ \exp(\bm{z}_i^\top \bm{z}'_q/\tau_c)}{\sum^{i \neq n}_i \exp(\bm{z}_i^\top \bm{z}'_n/\tau_c)} \right),
\end{equation}
where $\mathcal{N}_i$ represents the set of indices corresponding to images that share the same label as $\bm{x}_i$ within a batch $B$, and $\tau_c$ is a temperature parameter. Finally, the overall representation learning loss is:
\begin{equation}
	\mathcal{L}_{\text{rep}} = ( 1 - \lambda_{sim})\mathcal{L}^u_\text{rep} + \lambda\mathcal{L}^s_\text{rep}
\end{equation}

\textbf{2) Classifier Learning} aims to train a classifier that assigns labels to unlabeled data. Within the SimGCD framework, this objective is achieved through a parametric classifier refined via a self-distillation strategy, where the number of categories, denoted as $|\mathcal{Y}_u|$, is predetermined. Letting $K = |\mathcal{Y}_u|$, SimGCD initializes a set of parametric prototypes for each category, represented as $\mathcal{C} = \{\bm{c}_1, \bm{c}_2, \bm{c}_3, \ldots, \bm{c}_K\}$. Given a backbone network $f(\cdot)$, a soft label is obtained by applying softmax classification over these parametric prototypes:
\begin{equation}
{p}_i^k = \frac{\exp\left(\frac{1}{\tau_s} \left(\bm{h}_i/ \| \bm{h}_i \|_2\right)^{\top} \left(\bm{c}_k/ \left\|\bm{c}_k\right\|_2\right)\right)}{\sum_j \exp\left(\frac{1}{\tau_s} \left(\bm{h}_i/ \| \bm{h}_i \|_2\right)^{\top} \left(\bm{c}_j/ \left\|\bm{c}_j\right\|_2\right)\right)},  
\end{equation}
where ${\bm{h}_i}=f({\bm{x}_i})$ is the representation of ${\bm{x}_i}$ and $\tau_s$ is a temperature value. A soft label $\bm{q}'$ is similarly produced for $\bm{x}_i'$ with a sharper temperature $\tau_t$. The classification objectives are simply cross-entropy loss $ \mathcal{L}_{ce}{(\bm{q}',\bm{p})}=-\sum_k \bm{q}'^{(k)} \text{log} \bm{p}^{(k)} $ between the predictions and pseudo-labels or ground-truth labels. That is,

\begin{equation}
\mathcal{L}^{u}_{cls} = \frac{1}{|B|} \sum_{i \in B} \mathcal{L}_{ce}(\bm{q}'_i, \bm{p}_i) - \epsilon H(\bar{\bm{p}}),
\end{equation}
\begin{equation}
\quad \mathcal{L}^{s}_{cls} = \frac{1}{|B^l|} \sum_{i \in B^l} \mathcal{L}_{ce}(\bm{y}_i, \bm{p}_i) ,
\end{equation}
where $\bm{y}_i$ denotes the one-hot label of $\bm{x}_i$. SimGCD employs a mean-entropy maximization regularizer as part of the unsupervised objective. Specifically,  
$\bar{\bm{p}} = \frac{1}{2|B|} \sum_{i \in B} (\bm{p}_i + \bm{p}'_i)$ represents the mean prediction of a batch, and the entropy is defined as  
$H(\bar{\bm{p}}) = -\sum_k \bar{\bm{p}}^{(k)} \log \bar{\bm{p}}^{(k)}$.  
The classification objective is:
   \begin{equation}
    \mathcal{L}_{cls} = (1 - \lambda_{sim}) \mathcal{L}^{u}_{cls} + \lambda \mathcal{L}^{s}_{cls},  
\end{equation}
The overall objective of SimGCD is:
\begin{equation}
\mathcal{L}_{sim} = \mathcal{L}_{rep} + \mathcal{L}_{cls}.  
\end{equation}

\section{Experimental Setup}\label{sec:setup}

\subsection{The details of datasets}\label{sec:dataset}

\begin{table*}[t]
	\centering
	\begin{tabular}{c c c c c c c}
		\toprule
		Dataset & All(classes/samples) & {Old} labeled & {Old} Unlabeled & New & $\lambda$ & \(\tau\) \\
		\midrule
		CUB~\cite{welinder2010caltech} & 200/6k & 100/1.5k & 100/1.5k & 100/3k & 0.05 & 0.2 \\ 
		Stanford Cars~\cite{krause20133d} & 196/8.1k & 98/2.0k & 98/2.0k & 98/4.1k & 0.05 & 0.01 \\ 
		FGVC-Aircraft~\cite{maji2013fine} & 100/6.7k & 50/1.7k & 50/1.7k & 50/3.3k & 0.05 & 0.01\\
		CIFAR10~\cite{krizhevsky2009learning} & 10/50.0k & 5/12.5k & 5/12.5k & 5/25.0k & 0.05 & 0.1 \\
		CIFAR100~\cite{krizhevsky2009learning} & 100/50.0k & 80/20.0k & 80/12.5k & 20/17.5k & 0.05 & 0.1 \\
		ImageNet-100~\cite{deng2009imagenet} & 100/127.2k & 50/31.9k & 50/31.9k & 50/63.4k & 0.05 & 0.05 \\
		Herbarium-19~\cite{tan2019herbarium} & 683/34.2k & 341/8.9k & 341/8.9k & 342/16.4k & 0.05 & 1e-4 \\
		\bottomrule
	\end{tabular}
	\caption{Summary of datasets and training configurations.}
	\label{tab:datasets}
\end{table*} 

\begin{figure*}[t]  
	\centering  
	\includegraphics[width=1\textwidth]{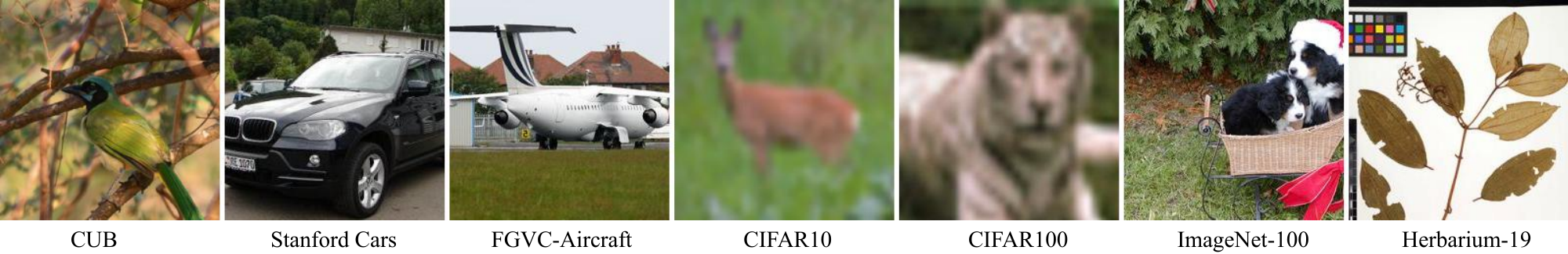}  
	\caption{Image examples from the used datasets.}  
	\label{fig:datasets}
	\vspace{-0.3cm}
\end{figure*} 

In this study, we validate the effectiveness of our method using three challenging fine-grained datasets from the Semantic Shift Benchmark~\cite{SSB2021}: \texttt{CUB}~\cite{welinder2010caltech}, \texttt{Stanford Cars}~\cite{krause20133d}, and \texttt{FGVC-Aircraft}~\cite{maji2013fine}. As illustrated in Figure~\ref{fig:datasets}, these datasets often contain complex background information. Following SimGCD~\cite{simGCD}, we partitioned each dataset into \emph{Known} and \emph{Unknown} categories, with each category representing 50\% of the total number of classes. Notably, 50\% of the samples in the \emph{Known} classes are unlabeled. To further assess the robustness of our method, we applied it to three generic classification datasets (\texttt{CIFAR10/100}~\cite{krizhevsky2009learning} and \texttt{ImageNet-100}~\cite{deng2009imagenet}), as well as the challenging large-scale fine-grained dataset \texttt{Herbarium-19}~\cite{tan2019herbarium}. As shown in Figure~\ref{fig:datasets}, the background interference in these datasets is relatively minimal. We employed the same partitioning strategy for these datasets, except for \texttt{CIFAR-100}, where 80\% of the classes were designated as \emph{Known} categories. Detailed information of datasets can be found in Table~\ref{tab:datasets}. 

\subsection{Implementation details} \label{sec:implement}
Following SimGCD~\cite{simGCD}, we trained all methods with a ViT-B/16 backbone~\cite{vit16b} pre-trained with DINO~\cite{caron2021emerging}. We use the output of AF with a dimension of 768 as the feature for an image and only fine-tune the last block of the backbone. We train with a batch size of 128 for 200 epochs with an initial learning rate of 0.1 decayed with a cosine schedule on each dataset. Aligning with~\cite{simGCD}, the balancing factor $\lambda_{sim}$ is set to 0.35, and the temperature values $\tau_u$, $\tau_c$ as 0.07, 1.0, respectively. For the classification objective, we set $\tau_s$ to 0.1, and $\tau_t$ is initialized to 0.07, then warmed up to 0.04 with a cosine schedule in the starting 30 epochs. For AF, the configurations of $\lambda$ and $\tau$ are provided in Table~\ref{tab:datasets}. All experiments are done with an NVIDIA GeForce RTX 4090 GPU.  
\section{Extended Discussions}\label{sec:discuss}

\subsection{The impact of AF on model attention}
\label{sec:App_attenMap}
To further investigate \emph{Distracted Attention} in the model across various data sets, we used the self-attention scores of the final ViT block to generate patch masks on both the \texttt{Stanford Cars} and \texttt{FGVC-Aircraft} datasets. As depicted in Figure \ref{fig:App_attenMap}, while the [CLS] tokens for labeled data consistently focus on key objects, those for unlabeled data, particularly from unknown category, exhibit pronounced associations with background regions. This unintended capture of extraneous information negatively impacts the quality of feature representations and, consequently, model performance. As can be observed from the comparison between different methods, AF significantly ameliorates the model's attention, enabling it to more effectively concentrate on the critical target regions. However, it is noteworthy that the extent of improvement varies across datasets due to differences in background complexity. As shown, \texttt{FGVC-Aircraft} predominantly features backgrounds such as airports or skies, which introduce minimal interference compared to the more cluttered and diverse backgrounds present in the \texttt{CUB} and \texttt{Stanford Cars}. This inherent characteristic of \texttt{FGVC-Aircraft} explains why the performance gains achieved through AF are less pronounced, compared to \texttt{CUB} and \texttt{Stanford Cars} (\textbf{Table 1 of Section 4.2}).

% It is noteworthy that the background scenarios in \texttt{FGVC-Aircraft} predominantly consist of airports or skies, which only introduce minimal interference. This is the primary reason why the improvement of AF on \texttt{FGVC-Aircraft} is significantly less pronounced compared to its performance on \texttt{CUB} and \texttt{Stanford Cars}. 
%  We observed an intriguing phenomenon: AF demonstrates more comprehensive attention to critical targets compared to SimGCD on \texttt{FGVC-Aircraft}. We attribute this phenomenon to the average aggregation following patch selection, which ensures that each retained patch contributes meaningfully to the overall representation. This process not only enhances the model's ability to focus on semantically significant regions but also improves the robustness of feature extraction. In contrast, SimGCD relies solely on the [CLS] token as its output, which progressively diminishes the model's ability to perceive the holistic structure of the target objects, ultimately leading to suboptimal performance. The relevant experimental results can be found in \textbf{Table 5} of \textbf{Section 4.3}. The results suggest that patch selection mechanism in AF, combined with average aggregation, effectively mitigates the loss of critical information, thereby enabling more precise attention allocation.

\begin{figure*}[ht]  
	\centering  
	\includegraphics[width=1\textwidth]{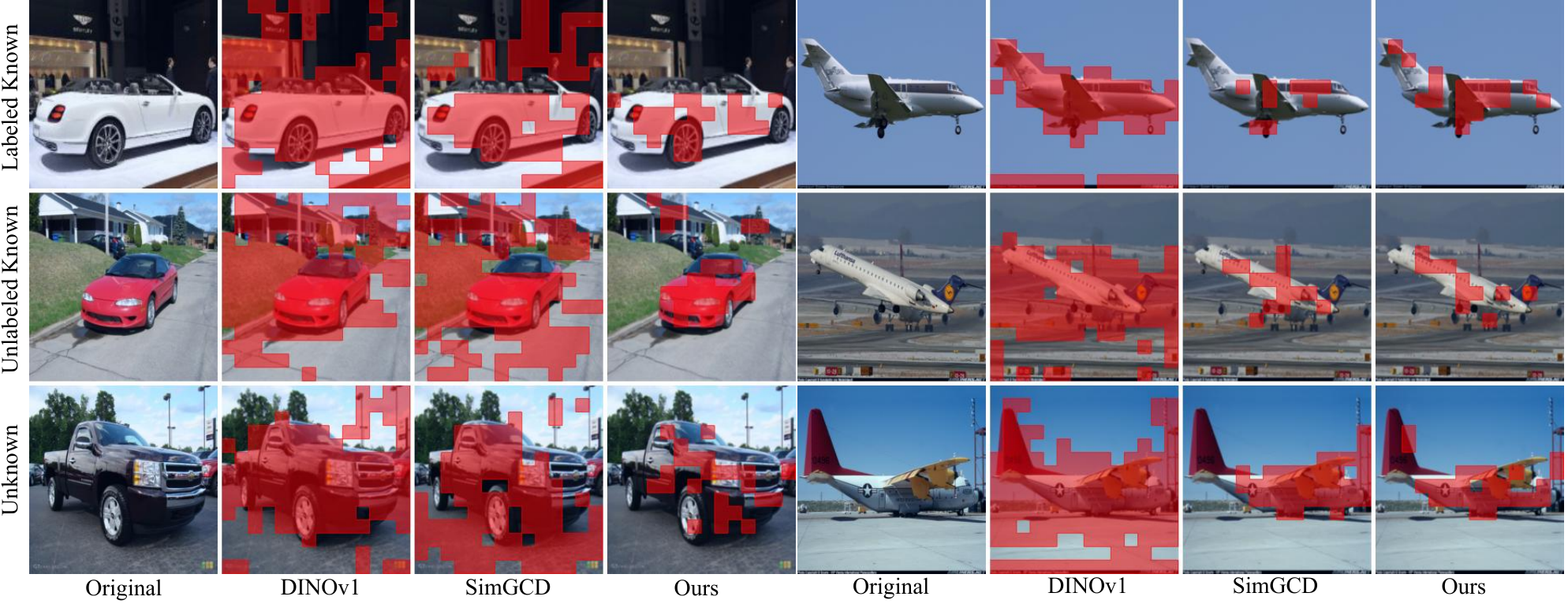}  
	\caption{The masks obtained by thresholding the self-attention maps to retain same percent of the total mass cross different methods.}  
	\label{fig:App_attenMap}
	\vspace{-0.3cm}
\end{figure*}

\subsection{Single-view TAP or Multi-view TAP?}\label{sec:single-view}

 During the training process of SimGCD+AF, each data point is augmented with two distinct views. And, TAP is applied to only one of these views. To further assess the potential benefits of a more comprehensive approach, we experimented with multi-view TAP, where TAP is applied to both augmented views simultaneously. As shown in Table~\ref{res:crop}, while multi-view TAP does offer some performance improvements, it also leads to a noticeable degradation in comparison to single-view TAP. We believe that this can be attributed to two primary factors. First, TAP can be viewed as a form of non-regular image cropping augmentation, where single-view TAP is particularly effective in helping the model focus on key objects or regions of interest. By pruning unnecessary tokens in a single view, the model is able to maintain critical information, thus improving its ability to extract meaningful features from the image. Second, multi-view TAP essentially forces the model to train without the potential interference of background information across both views. While this might seem beneficial in theory by reducing noise, it can inadvertently reduce the model's ability to generalize.

\begin{table}[ht]
	\footnotesize
	\centering
	\begin{tabular}{l p{0.3cm} p{0.3cm} p{0.3cm} p{0.3cm} p{0.3cm} p{0.3cm} p{0.3cm} p{0.3cm} p{0.3cm}}
		\toprule
		\multicolumn{1}{l}{\multirow{2}{*}{Datasets}} & \multicolumn{3}{c}{CUB} & \multicolumn{3}{c}{Stanford Cars} & \multicolumn{3}{c}{FGVC-Aircraft}  \\
		& All & Old & {New} & All & {Old} & {New} & All & {Old} & {New}   \\ 
		\midrule
		SimGCD  & 60.1 & 69.7 & 55.4 & 55.7 & 73.3 & 47.1 & 53.7 & 64.8 & 48.2  \\
		+AF(M-TAP) & 66.8 & 73.1 & 63.6 & 63.2 & 79.9 & 55.1 & 57.4 & 65.7 & 53.3 \\
		\rowcolor{lightgray}
		\textbf{+AF(S-TAP)} & {69.0} & 74.3 & {66.3} & {67.0} & {80.7} & {60.4} & {59.4} & {68.1} & {55.0} \\ 
		\bottomrule 
	\end{tabular}
	\caption{Investigation of \emph{Single-view Token Adaptive Pruning}. 'AF(M-TAP)' refers to a setting where TAP is applied to both augmented views simultaneously.} 
	\label{res:crop}
	\vspace{-0.3cm} 
\end{table}

\subsection{[CLS] token attention vs. AF}\label{sec:cls_attention}
To further demonstrate the effectiveness of AF, we utilize the attention weights between the [CLS] token and individual patches as the scores in AF, while employing the same strategy for pruning. The experimental results, as presented in Table~\ref{res:clsToken}, reveal that constraining the interaction between the [CLS] token and the irrelevant patches to a certain extent indeed enhances model performance. This improvement underscores the utility of refining the model's attention by mitigating the influence of task-irrelevant regions. However, it is particularly noteworthy that accuracy for \emph{Old} category on \texttt{FGVC-Aircraft} exhibits a pronounced decline. This phenomenon suggests that the attention weights derived solely from the internal interactions between the [CLS] token and other patches are inadequate to guarantee that the model consistently attends to the correct key target regions. Such an outcome highlights the limitations of relying exclusively on intrinsic attention mechanism without additional guidance or constraints. Collectively, these findings not only underscore the generalizability and robustness of AF in diverse datasets, but also emphasize the necessity of incorporating more sophisticated strategies to ensure precise attention allocation in complex visual recognition tasks.

 \begin{table}[h]
 	\footnotesize
 	\centering
 	\begin{tabular}{l p{0.3cm} p{0.3cm} p{0.3cm} p{0.3cm} p{0.3cm} p{0.3cm} p{0.3cm} p{0.3cm} p{0.3cm}}
 		\toprule
 		\multicolumn{1}{l}{\multirow{2}{*}{Datasets}} & \multicolumn{3}{c}{CUB} & \multicolumn{3}{c}{Stanford Cars} & \multicolumn{3}{c}{FGVC-Aircraft}  \\
 		& All & Old & {New} & All & {Old} & {New} & All & {Old} & {New}   \\ 
 		\midrule
 		SimGCD  & 60.1 & 69.7 & 55.4 & 55.7 & 73.3 & 47.1 & 53.7 & 64.8 & 48.2  \\
 		+([CLS] Atten)  & 63.9 & 72.2 & 59.8 & 62.3 & 77.4 & 55.1 & 54.9 & 58.2 & 53.3  \\
 		\rowcolor{lightgray}
 		\textbf{+AF} & {69.0} & 74.3 & {66.3} & {67.0} & {80.7} & {60.4} & {59.4} & {68.1} & {55.0} \\ 
 		\bottomrule 
 	\end{tabular}
 	\caption{Investigation of \emph{[CLS] Token Attention}. 'AF([CLS] Atten)' refers to using the attention weights between the [CLS] Token and patches as patch scores.}
 		\label{res:clsToken}
 		\vspace{-0.3cm}
 	\end{table}

\begin{figure*}[ht]  
	\centering  
	\includegraphics[width=1\textwidth]{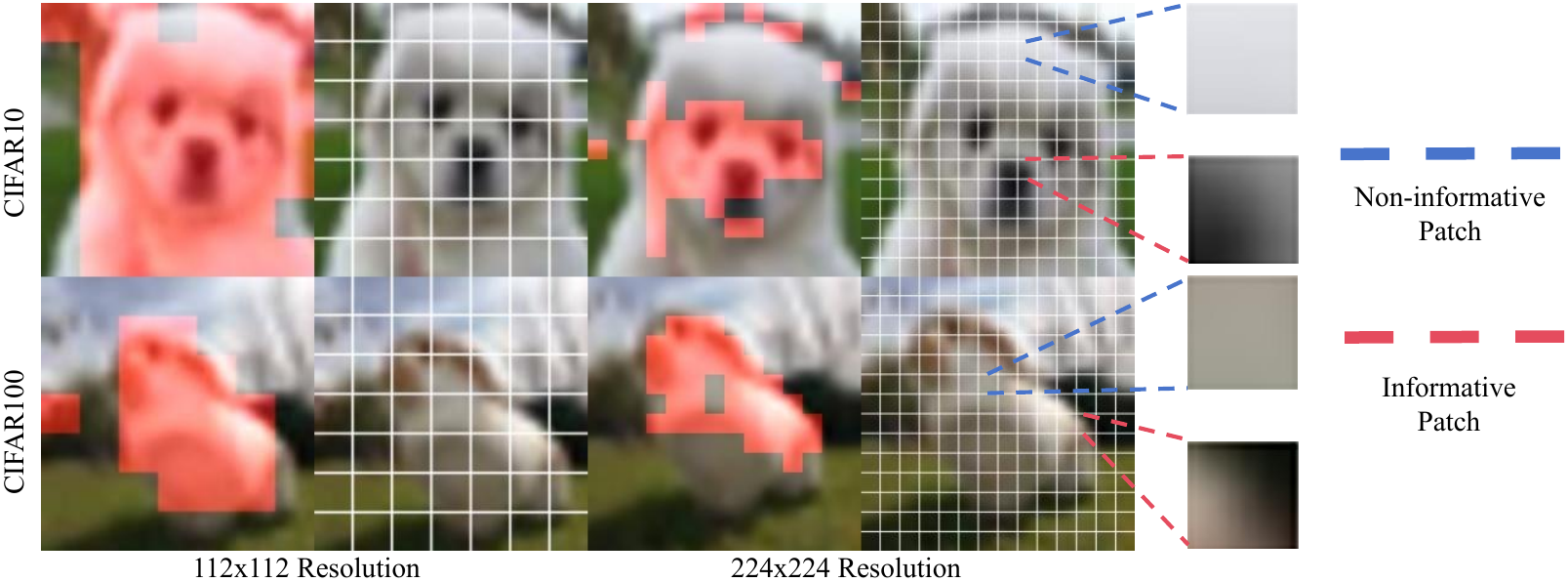}  
	\caption{The partitions of input images with the same patch size under different resolutions.}  
	\label{fig:resolution}
	\vspace{-0.3cm}
\end{figure*} 

\subsection{The impact of resolution}
\label{sec:resolution}  
Our empirical evaluations reveal that Attention Focusing (AF) demonstrates limited performance improvements on \texttt{CIFAR10/100}, prompting a systematic investigation into its constraints. To this end, we conducted controlled experiments involving resolution scaling of input images. As illustrated in Figure~\ref{fig:resolution}, original 32×32 pixel images were upsampled to target resolutions of 112×112 and 224×224, followed by uniform patch selection strategies under AF. Notably, a critical phenomenon emerged when maintaining consistent patch size across resolutions: Some internal patches of the target contain less information in high-resolution input images. For instance, the blue-dashed area in Figure~\ref{fig:resolution} highlights a region devoid of meaningful texture, which the TIME module assigns a low significance score due to insufficient structural information. This selection bias induces cascading effects, including (1) loss of global object-related information during representation reconstruction and (2) suboptimal feature extraction due to discarding foundational constituent patches. Quantitative experiments in Table~\ref{res:resolution} corroborates these observations: 224×224 resolution fails to achieve remarkable performance improvements, even exhibiting performance degradation on \texttt{CIFAR100}, whereas adopting 112×112 resolution not only yields significant performance gains but also substantially reduces computational cost by over 70\%, with FLOPs decreasing from 16.87G to 4.7G.

This finding establishes a critical implementation protocol for AF: Processing original low-resolution images through moderate resolution scaling achieves synergistic optimization of model performance and computational efficiency by balancing information integrality with operational cost constraints.

\begin{table}[t]
	\footnotesize
	\centering
	\begin{tabular}{l p{0.27cm} p{0.27cm} p{0.27cm} p{0.27cm} p{0.27cm} p{0.27cm} l }
		\toprule
		\multicolumn{1}{l}{\multirow{2}{*}{Datasets}} & \multicolumn{3}{c}{CIFAR10} & \multicolumn{3}{c}{CIFAR100} & \multicolumn{1}{l}{\multirow{2}{*}{FLOPs}} \\
		& All & {Old} & {New} & All & {Old} & {New}    \\ 
		\midrule
		SimGCD~\cite{simGCD}  & 97.1 & 95.1 & 98.1 & 80.1 & 81.2 & 77.8 & 16.87G\\
		SimGCD+AF(224x224) &  97.4 & 95.7 & 98.3 & 79.8 & 83.5 & 72.4 & 18.32G\\
		\rowcolor{lightgray} 
		SimGCD+AF(112x112) & 97.8 & 95.9 & 98.8 & 82.2 & 85.0 & 76.5 & \; 4.7G\\ 
		\bottomrule 
		
	\end{tabular}
	\caption{Comparison with different resolutions.}
	\label{res:resolution}
	\vspace{-0.3cm}
\end{table}

\subsection{Class Token or Aggregation Token?}
\label{sec:clstoken}
In AF, we compute the average of all remaining tokens, including the [CLS] token, to represent the image feature, which serves as the output of the backbone. The rationale behind this approach is that the remaining tokens are considered key patches that contain critical information about the object. In contrast, a common practice is to use only the [CLS] token as the image representation. As shown in Table~\ref{res:clsToken}, this approach results in a significant drop in performance. We believe the primary cause of this decline is that applying the self-attention mechanism solely in the final block prevents the [CLS] token from effectively aggregating information from the diverse patches throughout the image.

\begin{table}[h]
    \hspace{-0.5cm}
    \footnotesize
    \centering
\begin{tabular}{l p{0.3cm} p{0.3cm} p{0.3cm} p{0.3cm} p{0.3cm} p{0.3cm} p{0.3cm} p{0.3cm} p{0.3cm}}
\toprule
\multicolumn{1}{l}{\multirow{2}{*}{Datasets}} & \multicolumn{3}{c}{CUB} & \multicolumn{3}{c}{Stanford Cars} & \multicolumn{3}{c}{FGVC-Aircraft}  \\
       & All & Old & {New} & All & {Old} & {New} & All & {Old} & {New}   \\ 
       \midrule
         SimGCD  & 60.1 & 69.7 & 55.4 & 55.7 & 73.3 & 47.1 & 53.7 & 64.8 & 48.2  \\
         +AF([CLS]) & 65.2 & 69.5 & 63.1 & 56.2 & 75.9 & 46.6 & 54.6 & 65.6 & 49.1 \\
         \rowcolor{lightgray}
         \textbf{+AF} & {69.0} & 74.3 & {66.3} & {67.0} & {80.7} & {60.4} & {59.4} & {68.1} & {55.0} \\ 
         \bottomrule 
    \end{tabular}
    \caption{Investigation of \emph{Token Aggregation}. 'AF([CLS])' refers to a setting where the [CLS] token is used as the output of the backbone.}
    \label{res:clsToken}
    \vspace{-0.6cm}
\end{table}
\vspace{4pt}

\subsection{Computational efficiency of AF}
\label{sec:efficiency}
To further validate the lightweight characteristics of AF module, we conducted quantitative comparisons during both training and inference phases. As illustrated in Table~\ref{res:efficiency}, while the parameter exhibits a more substantial increase during the training phase, the increase becomes negligible during inference —- each TIME module requires only a single vector for computation. Notably, despite the increased training parameters, the additional computational overhead remains marginal, with only a modest prolongation in training time consumption. Similarly, the testing time demonstrates merely a slight increment. These results underscore that the AF module achieves enhanced functionality without substantially compromising computational efficiency. The minimal impact on inference phase makes it particularly suitable for deployment in resource-constrained environments. 

\begin{table}[t]
	\footnotesize
	\centering
	\begin{tabular}{l c c c c }
		\toprule
		\multicolumn{1}{l}{\multirow{2}{*}{Method}} & \multicolumn{2}{c}{Parameter quantity} & \multicolumn{2}{c}{Time consumption} \\
		& Training & Testing & Training & Testing    \\ 
		\midrule
            SimGCD &    81.82M  & 81.82M & 18.875s & 8s \\
            SimGCD+AF & 132.21M & 81.83M & 21.125s & 10s\\
		\bottomrule 
		
	\end{tabular}
	\caption{Quantitative comparison of parameter quantities and time consumption for training and testing phases.}
	\label{res:efficiency}
	\vspace{-0.3cm}
\end{table}

\subsection{Parameter analysis}
\label{sec:parameter}

\noindent\emph{1) Hyperparameter} $\tau$

For $\tau$, we maintain $\lambda=0.05$, while varying $\tau$ with a same interval. As shown in Figure \ref{fig:param}, it is evident that $\tau$ can yield significant performance improvements within a specific range. However, the influence of $\tau$ on model performance is particularly pronounced, as it directly governs the extent of redundant information pruning. When $\tau$ is excessively large or small, it leads to over-pruning and under-pruning, respectively. Over-pruning results in the loss of global information, while under-pruning retains excessive redundancy, both of which adversely affect the model's performance. Furthermore, the inherent variability of key target regions across images, influenced by differences in object scale, spatial distribution, and background complexity, makes a fixed pruning amount suboptimal. This limitation is empirically demonstrated in \textbf{Table 7} of \textbf{Section 4.3}, where fixed pruning strategies underperform compared to adaptive approaches. Such variability highlights the need for a more flexible pruning framework that can dynamically adjust to the unique image.

% Moreover, the inherent variability of key target regions in images makes a fixed pruning amount a suboptimal choice(\textbf{Table 7 in Section 4.3}). As can be seen from the figure, as long as $\tau$ remains at a relatively low value (i.e., redundant information is pruned), the model performance improves. This further validates that our proposed AF is fundamentally a sound and effective approach.

\vspace{0.2cm}
\noindent\emph{2) Hyperparameter} $\lambda$

For $\lambda$, we maintain $\tau$ as the pre-set value for the corresponding dataset, while varying  within the set $\lambda$ = \{0.01, 0.03, 0.05, 0.07, 0.1\}. As shown in Figure~\ref{fig:param}, it can be observed that the performance of AF declines when $\lambda \leq 0.03$. We attribute this phenomenon to the excessively low auxiliary loss, which diminishes the model's ability to prune redundant information. This reduction in pruning capacity leads to a lower pruning rate, resulting in the retention of excessive irrelevant features and, consequently, a degradation in representation. Conversely, when the loss is excessively high, the pruning rate of AF becomes overly aggressive, leading to incomplete image representations due to the excessive removal of critical information. These observations reveal a clear relationship between the auxiliary loss and the pruning rate: the loss function directly influences the model's pruning behavior by controlling the trade-off between retaining relevant features and eliminating redundancy. Despite these variations, AF consistently achieves significant performance improvements across different $\lambda$, demonstrating its robustness and effectiveness in enhancing image representation.

\begin{figure}[t]
	\hspace*{-0.4cm}
	\includegraphics[height=2.1cm]{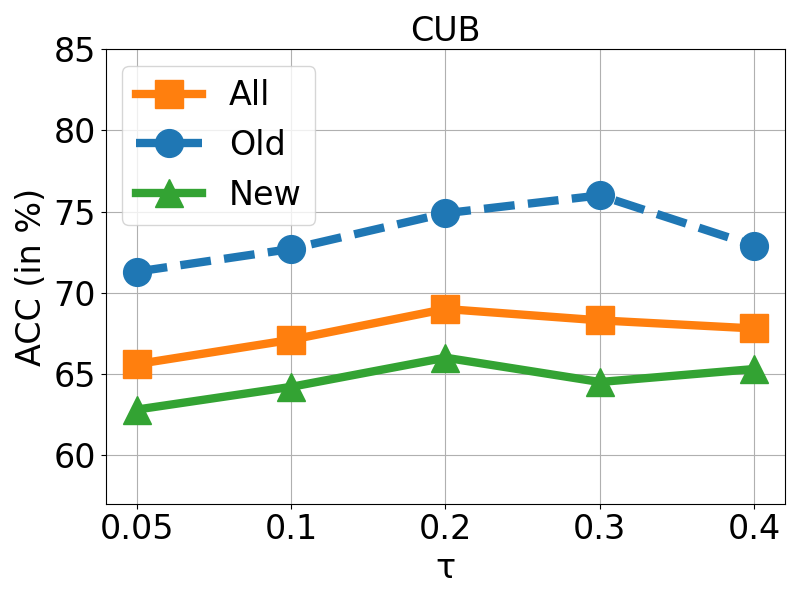}
	\includegraphics[height=2.1cm]{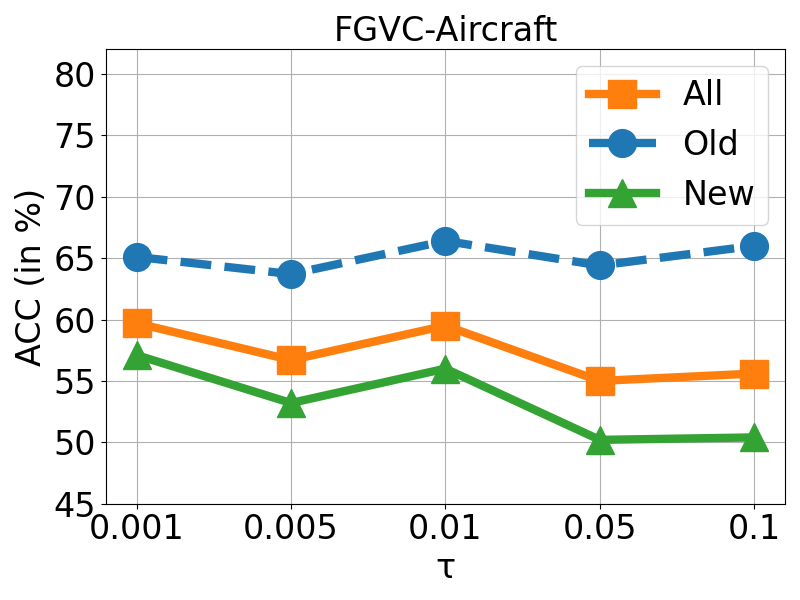}
	\includegraphics[height=2.1cm]{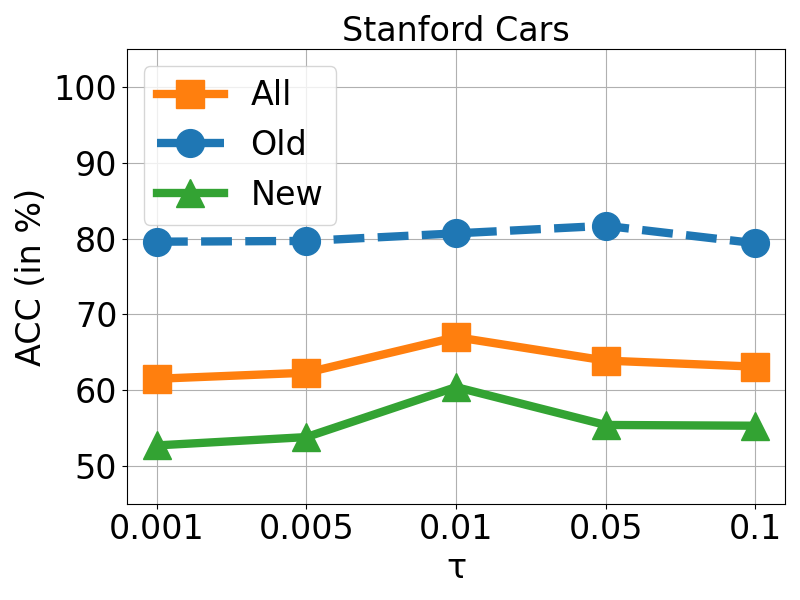}
	\hspace*{-0.35cm}
	\includegraphics[height=2.1cm]{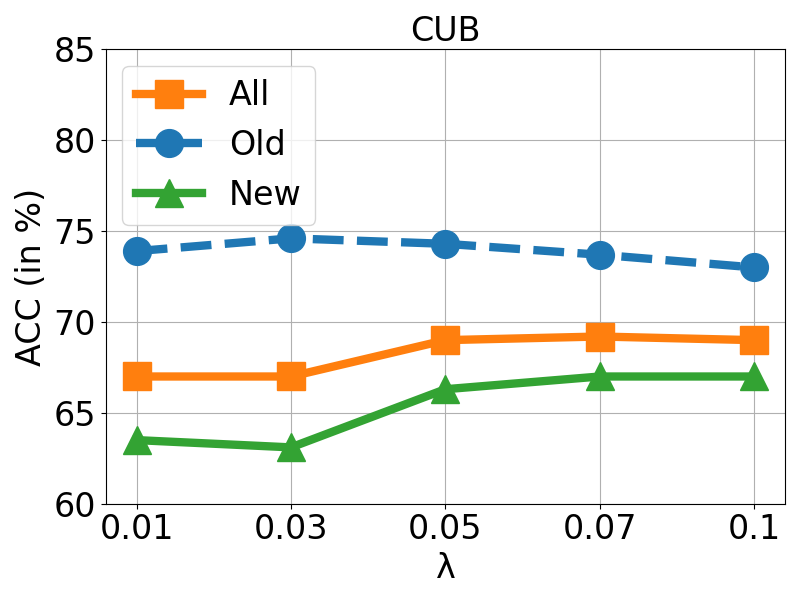}
	\includegraphics[height=2.1cm]{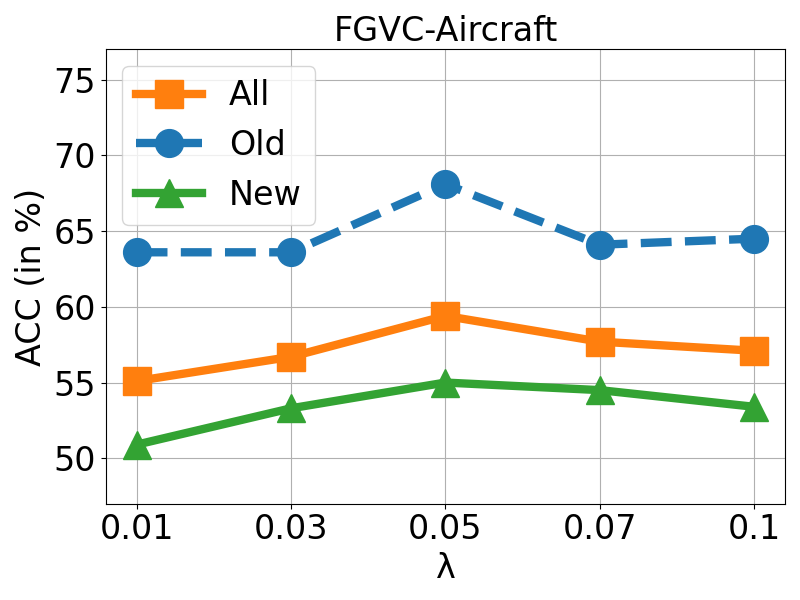}
	\includegraphics[height=2.1cm]{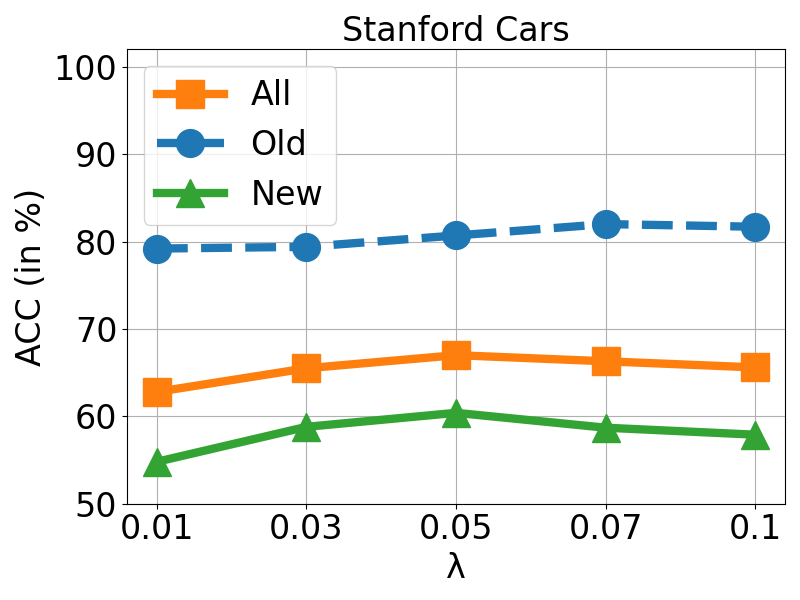}
	\caption{Investigation of the parameter $\lambda$ and  $\tau$.}
	\label{fig:param}
\end{figure}

% As shown Figure~\ref{fig:param}, it can be observed that the performance of AF declines when $\lambda \leq 0.03$. Through our experiments, we attribute this phenomenon to the excessively low auxiliary loss, which results in a reduced pruning rate in AF. Consequently, an excessive amount of redundant information is retained, adversely affecting the accuracy of image representation. Conversely, when the loss is excessively high, the pruning rate of AF becomes overly aggressive, leading to incomplete image representations due to the excessive removal of critical information. Despite these variations, AF consistently achieves significant performance improvements across different $\lambda$, demonstrating its robustness and effectiveness in enhancing image representation.

% {\small
% \bibliographystyle{ieeenat_fullname}
% \bibliography{reference}
% }

\end{document}